\documentclass[letterpaper,twocolumn,10pt]{article}
\usepackage{times}
\usepackage{comment,soul,times}
\usepackage{graphicx,tabulary}
\usepackage{amsmath,amssymb,mathrsfs}
\usepackage{breakurl,authblk} 
\usepackage[breaklinks]{hyperref}
\usepackage{algorithm,algpseudocode}
\usepackage{usenix,epsfig,endnotes,color}
\usepackage{subcaption}
\usepackage[font={small}]{caption}
\usepackage{setspace}
\usepackage{nopageno}
\usepackage{color}
\usepackage{amsmath}
\usepackage{amssymb}
\usepackage{multirow, varwidth}
\usepackage{stfloats}
\usepackage{verbatim}
\makeatletter
\newif\if@restonecol
\makeatother
\let\oldnl\nl
\newcommand{\nonl}{\renewcommand{\nl}{\let\nl\oldnl}}
\usepackage{xr}
\usepackage[amsmath,thmmarks]{ntheorem}

\newcolumntype{K}[1]{>{\centering\arraybackslash}m{#1}}
\newcolumntype{L}[1]{>{\raggedright\arraybackslash}p{#1}}

\theorembodyfont{\normalfont}
\theoremseparator{.}

\theoremstyle{nonumberplain}

\DeclareRobustCommand\onedot{\futurelet\@let@token\@onedot}
\def\onedot{. }

\def\etal{\emph{et al}\onedot}

\DeclareMathAlphabet{\pazocal}{OMS}{zplm}{m}{n}
\newcommand{\Lb}{\pazocal{L}}

\begin{document}
\newcommand{\FIXME}[1]{{\color{red}{\small\bf\sf [#1]}}}
\newcommand{\hao}[1]{{\color{red}{\small\bf\sf [Hao: #1]}}}
\newcommand{\haox}[1]{{\color{red}{\sf \bf [Hao: #1]}}}
\date{}

\title{\Large \bf Poseidon: An Efficient Communication Architecture for Distributed Deep Learning on GPU Clusters}
\author{
{\rm Hao Zhang$^{1,2}$, Zeyu Zheng$^{2}$, Shizhen Xu$^{1}$, Wei Dai$^{1,2}$, Qirong Ho$^{2}$, Xiaodan Liang$^{1}$, \vspace{-10pt}}\\
{\rm Zhiting Hu$^{1, 2}$, Jinliang Wei$^{1}$, Pengtao Xie$^{1,2}$, Eric P. Xing$^{2}$ \vspace{3pt}} \\
{\small Carnegie Mellon University$^{1}$, Petuum Inc.$^2$}
} 

\maketitle


\subsection*{Abstract}

Deep learning models can take weeks to train on a single GPU-equipped machine, necessitating scaling out DL training to a GPU-cluster. However, current distributed DL implementations can scale poorly due to substantial parameter synchronization over the network, because the high throughput of GPUs allows more data batches to be processed per unit time than CPUs, leading to more frequent network synchronization. We present Poseidon, an efficient communication architecture for distributed DL on GPUs. Poseidon exploits the layered model structures in DL programs to overlap communication and computation, reducing bursty network communication. Moreover, Poseidon uses a hybrid communication scheme that optimizes the number of bytes required to synchronize each layer, according to layer properties and the number of machines. 
We show that Poseidon is applicable to different DL frameworks by plugging Poseidon into Caffe and TensorFlow. We show that Poseidon enables Caffe and TensorFlow to achieve 15.5x speed-up on 16 single-GPU machines, even with limited bandwidth (10GbE) and the challenging VGG19-22K network for image classification. Moreover, Poseidon-enabled TensorFlow achieves 31.5x speed-up with 32 single-GPU machines on Inception-V3, a 50\% improvement over the open-source TensorFlow (20x speed-up). 



\vspace{-10pt}
\section{Introduction}
\vspace{-5pt}

Deep learning (DL) is a class of machine learning (ML) approaches that has achieved notable success across a wide spectrum of tasks, including speech recognition~\cite{Deng:2013:ICASSP}, visual recognition~\cite{Yan:2015:HDCNN,yan2016automatic} and language understanding~\cite{Mikolov:2013:ICLRW,liang2017recurrent}. These DL models exhibit a high degree of model complexity, with many parameters in deeply layered structures that usually take days to weeks to train on a GPU-equipped machine. 
The high computational cost of DL programs on large-scale data necessitates the training on distributed GPU cluster in order to keep the training time acceptable.



DL software such as TensorFlow~\cite{abadi2016tensorflow} and Caffe~\cite{Jia:2014:MM} allow practitioners to easily experiment with DL models on a single machine. 
However, their distributed implementations can scale poorly for larger models. For example, we find that on the VGG19-22K network (229M parameters), open-source TensorFlow on 32 machines can be slower than single machine (Section~\ref{sec:evaluation:scalability}). This observation underlines the challenge of scaling DL on GPU clusters: the high computational throughput of GPUs allows more data batches to be processed per minute (than CPUs), leading to more frequent network synchronization that grows with the number of machines. Existing communication strategies, such as parameter servers (PS) for ML~\cite{Wei:2015:SoCC,li2014scaling}, can be overwhelmed by the high volume of communication~\cite{cui2016geeps}.
Moreover, despite the increasing availability of faster network interfaces such as Infiniband or 40GbE Ethernet, GPUs have continued to grow rapidly in computational power, and continued to produce parameter updates faster than can be naively synchronized over the network. For instance, on a 16-machine cluster with 40GbE Ethernet and one Titan X GPU per machine, updates from the VGG19-22K model will bottleneck the network, so that only an 8x speedup over a single machine is achieved (Section~\ref{sec:evaluation:scalability}).

These scalability limitations in distributed DL stem from at least two causes: (1) the gradient updates to be communicated are very large matrices, which quickly saturate network bandwidth; (2) the iterative nature of DL algorithms causes the updates to be transmitted in bursts (at the end of an iteration or batch of data), with significant periods of low network usage in between. We propose that a solution to these two problems should exploit the structure of DL algorithms on two levels: on one hand, it should identify ways in which the matrix updates can be separated from each other, and then schedule them in a way that avoids bursty network traffic. On the other hand, the solution should also exploit the structure of the matrix updates themselves, and wherever possible, reduce their size and thus the overall load on the network. For such a solution to be relevant to practitioners (who may have strong preferences for particular frameworks), we would prefer not to exploit specific traits of TensorFlow's or Caffe's design, but should strive to be relevant to as many existing frameworks as possible.

With this motivation, we design Poseidon, an efficient communication architecture for data-parallel DL on distributed GPUs. Poseidon exploits the sequential layer-by-layer structure in DL programs, finding independent GPU computation operations and network communication operations in the training algorithm, so that they can be scheduled together to reduce bursty network communication. Moreover, Poseidon implements a hybrid communication scheme that accounts for each DL program layer's mathematical properties as well as the cluster configuration, in order to compute the network cost of different communication methods, and select the cheapest one -- currently, Poseidon implements and supports a parameter server scheme~\cite{Wei:2015:SoCC} that is well-suited to small matrices, and a sufficient factor broadcasting scheme~\cite{Xie:2015:arXiv} that performs well on large matrices. We focus on synchronous parallel training which is shown to yield faster convergence compared with asynchronous training in distributed DL (as measured by wall clock time) on GPUs~\cite{cui2016geeps,chen2016revisiting}.
Unless otherwise specified, our discussion in this paper assumes synchronous replication of model parameters in each training iteration, although we note that Poseidon's design can easily be applied to asynchronous or bounded-asynchronous consistency models~\cite{Ho:2013:NIPS,essp}.

To demonstrate Poseidon's applicability to multiple DL frameworks, we implement it into two different DL frameworks: Caffe and TensorFlow, and show that Poseidon allows them to scale almost-linearly in algorithm throughput with additional machines, while incurring little additional overhead even in the single machine setting. For distributed execution, with 40GbE network bandwidth available, Poseidon consistently delivers near-linear increases in throughput across various models and engines: 31.5x speedup on training the Inception-V3 network using TensorFlow engine on 32 nodes, which improves 50\% upon the original TensorFlow (20x); when training a 229M parameter network (VGG19-22K), Poseidon still achieves near-linear speedup (30x on 32 nodes) using both Caffe and TensorFlow engines, while distributed TensorFlow sometimes experiences negative~\cite{zhang2015poseidon} scaling with additional machines. Our experiments also confirm that Poseidon successfully alleviates network communication bottlenecks, by reducing the required bandwidth for parallelizing large models. For example, when training VGG19-22K under limited bandwidth (10GbE), in contrast to a PS-based parallelization which only achieves 4x speedup with 16 machines, Poseidon effectively reduces the communication overheads by automatically specializing the best communication method for each layer, and is able to keep linearly scaling with throughput.
Compared to other communication reduction methods~\cite{Chilimbi:2014:OSDI,yu2014introduction}, Poseidon demonstrates either systems advantages (increased algorithm throughput) or statistical advantages (fewer algorithm steps or iterations to reach a fixed termination criteria). Poseidon does not suffer much from imbalanced communication loads, which we found to be the case when using the sufficient factor strategy used in Project Adam~\cite{Chilimbi:2014:OSDI}. Poseidon also guarantees that the number of algorithm steps to reach termination remains unchanged, unlike the 1-bit quantization strategy used in CNTK~\cite{yu2014introduction} which is approximate and can hurt statistical performance in some applications. 

The rest of the paper is organized as follows. Section~\ref{sec:background} motivates Poseidon with introduction on large-scale DL, parameter servers and sufficient factor broadcasting. Section~\ref{sec:design} and section~\ref{sec:implementation} elaborates Poseidon's design and implementation, respectively. Section~\ref{sec:evaluation} evaluates Poseidon by training different models over multiple datasets, including comparisons to state-of-the-art GPU-based distributed DL systems.
Section~\ref{sec:relatedwork} discusses related works and section~\ref{sec:conclusion} concludes.

\vspace{-15pt}
\section{Large-scale Deep Learning}
\vspace{-8pt}
\label{sec:background}
In this section, we formulate the DL training as an iterative-convergent algorithm, and describe parameter server (PS) and sufficient factor broadcasting (SFB) for parallelizing such computation on clusters.
\vspace{-12pt}
\subsection{Distributed Deep Learning}
\begin{figure}[]
\centering
\includegraphics[width=0.85\columnwidth]{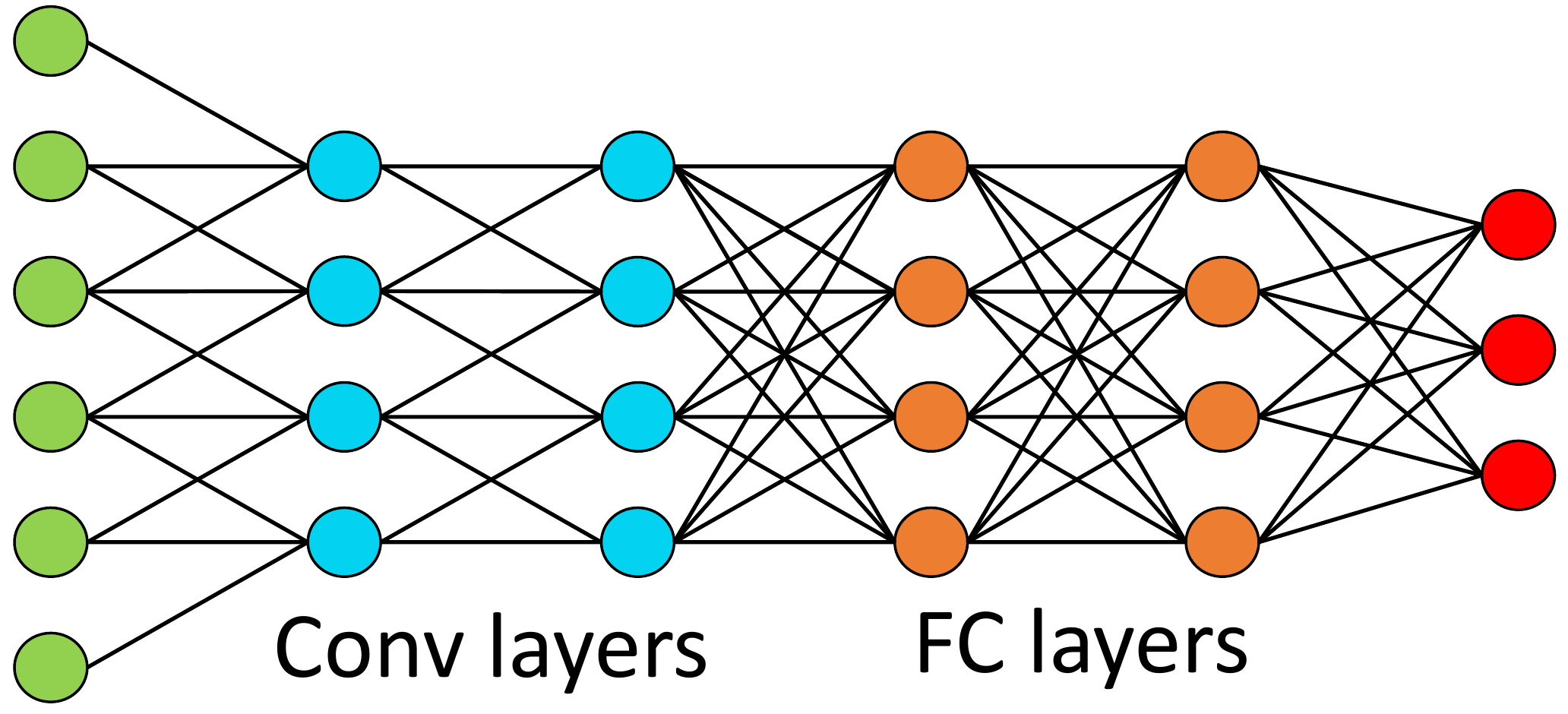}
\vspace{-8pt}
\caption{\small A convolutional neural network with 6 layers.}
\vspace{-12pt}
\label{fig:nn}
\end{figure}

\vspace{-5pt}
DL programs are distinguished from other ML programs mainly by their use of neural networks (NNs), a family of hierarchical models containing many layers, from as few as 5-10~\cite{Krizhevsky:2012:NIPS} to as many as 100s~\cite{he2015deep}. 
Figure \ref{fig:nn} illustrates a neural network with 6 layers.
The first layer (green) is an input layer that reads data in application-specific formats, e.g., raw pixels if it is trained to classify images.
The input layer is connected to a sequence of intermediate layers (cyan, orange), each of which consists of a few neurons, where each neuron applies a function transformation $f$ on its input and produces an output. A vector output is obtained by concatenating the output of all neurons from a layer. By stacking multiple intermediate layers, the NN can transform raw input data one layer at a time, first into a series of intermediate representations, and finally into the desired output or prediction (red).
DL programmers usually need to specify the computation of a layer by defining two properties of its neurons. The first is the transformation function $f(W, x)$, where $x$ is the input to the neuron, and $W$ is an \emph{optional trainable} parameter.
The other is the connectivity that determines how the neuron should be connected to its adjacent layer. For instance, a convolutional neural network has two types of neuron: convolutional (CONV) neuron (cyan) that are only locally connected to a subset of neurons in its previous layer, and fully-connected (FC) neurons (orange).
\begin{figure}[]
\small
\centering
\includegraphics[width=\linewidth]{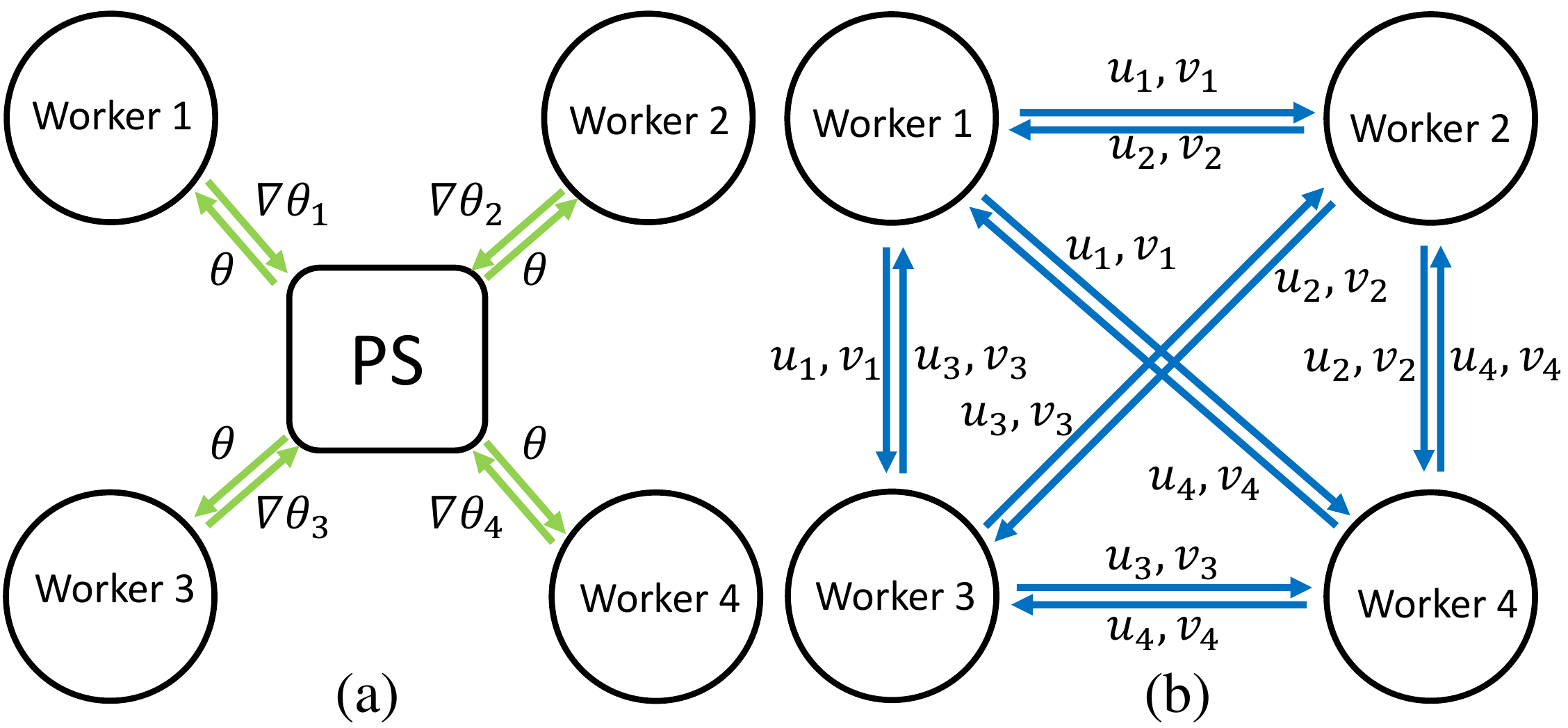}
        \vspace{-20pt}
        \caption{An illustration of (a) the parameter server and (b) sufficient factor broadcasting for distributed ML.}
\label{fig:pssfb}
\vspace{-18pt}
\end{figure}
Most NNs need to be trained with data to give accurate predictions. Stochastic gradient descent (SGD) and backpropagation are commonly employed to train NNs iteratively -- each iteration performs a feed forward (FF) pass followed with a backpropagation (BP) pass. In the FF pass, the network takes a training sample as input, forwards from its input layer to output layer to produce a prediction. A loss function is defined to evaluate the prediction error, which is then backpropagated through the network in reverse, during which the network parameters are updated by their gradients towards where the error would decrease. After repeating a sufficient number of passes, the network will usually converge to some state where the loss is close to a minima, and the training is then terminated.
In a mathematical form, given data $D$ and a loss function $\Lb$, fitting the parameters $\theta$ of a NN can be formulated as an \emph{iterative-convergent} algorithm that repeatedly executing the update equation 
\begin{equation}
\vspace{-3pt}
\label{eq:iter_conv_cnn}
\theta^{(t)} =  \theta^{(t-1)} +  \epsilon \cdot \nabla_{\Lb} (\theta^{(t-1)}, D^{(t)})
\vspace{-3pt}
\end{equation}
until $\theta$ reaches some stopping criteria, where $t$ denotes the iteration. The update function $\nabla_{\Lb}$ calculates the gradients of $\Lb$ over current data $D_i (D_i \in D)$. The gradients are then scaled by a learning rate $\epsilon$ and applied on $\theta$ as updates. As the gradients are additive over data samples $i$, i.e., $\theta^{(t)} = \theta^{(t-1)} + \epsilon \cdot \sum_i \nabla_{\Lb} (\theta^{(t-1)}, D_i)$, for efficiency, we usually feed a batch of training samples $D^{(t)} (D^{(t)} \subset D)$ at each training iteration $t$, as in Eq.\ref{eq:iter_conv_cnn}.

In large-scale deep learning, data $D$ are usually too large to process on a single machine in acceptable time. To speedup the training,
we usually resort to \emph{data parallelism}, a parallelization strategy that partitions the data $D$ and distributes to a cluster of computational worker machines (indexed by $p=1,\cdots,P$), as illustrated in Figure \ref{fig:pssfb}. At each iteration $t$, every worker fetches a batch $D_p^{(t)}$ from its data partition and computes the gradients $\nabla_{\Lb} (\theta^{(t)}, D_p^{(t)})$.
Gradients from all workers are then aggregated and applied to update $\theta^{(t)}$ to $\theta^{(t+1)}$ following
\vspace{-10pt}
\begin{equation}
\label{eq:iter_conv_data_paral}
\theta^{(t + 1)} =  \theta^{(t)} +  \epsilon \sum_{p=1}^P \nabla_{\Lb} (\theta^{(t)}, D_p^{(t)})
\vspace{-7pt}
\end{equation}
Data-parallelism allows data to be locally partitioned to each worker, which is advantageous for large datasets. It however requires every worker to have read and write access to the shared model parameters $\theta$, which causes communication among workers; this shared access can be provided by a parameter server architecture~\cite{Wei:2015:SoCC,Chilimbi:2014:OSDI} (Figure~\ref{fig:pssfb}a) or a peer-to-peer broadcasting architecture~\cite{Xie:2015:arXiv} (Figure~\ref{fig:pssfb}b), both are designed for general-purpose data-parallel ML programs on CPUs.

\noindent \textbf{Parameter Server.}
A parameter server (PS) is a distributed shared memory system that provides systematic abstraction of iterative-convergent algorithms in data-parallel distributed ML. Typically, PS enables each worker to access the global model parameters $\theta$ via network communications following the client-server scheme. DL can be trivially parallelized over distributed workers using PS with the following 3 steps: (1) Each worker computes the gradients ($\nabla_{\Lb}$) on their own data partition and send them to remote servers; (2) servers receive the updates and apply ($+$) them on globally shared parameters; (3) a consistency scheme coordinates the synchronization among servers and workers (Figure~\ref{fig:pssfb}a).

\noindent \textbf{Sufficient Factor Broadcasting.}
Many ML models represent their parameters $\theta$ as matrices. 
For example, fully-connected NNs, when trained using SGD, their gradient $\nabla \theta$ over a training sample is a rank-1 matrix, which can be cast as the outer product of two vectors $u, v$: $\nabla \theta = uv^\top$, where $u$ and $v$ are called \textit{sufficient factors} (SFs). Sufficient factor broadcasting (SFB)~\cite{Xie:2015:arXiv} is designed to parallelize these models by broadcasting SFs among workers and then reconstructing the gradient matrices $\nabla \theta$ using $u, v$ locally. SFB presents three key differences from PS: (1) SFB uses a P2P communication strategy that transmits SFs instead of full matrices. (2) Unlike gradients, SFs are not additive over training samples, i.e., the number of SFs needed to be transmitted grows linearly with the number of data samples (not data batches); (3) the overall communication overheads of SFB increase quadratically with the number of workers.  

\vspace{-10pt}
\subsection{Parallel DL on Distributed GPUs}
\vspace{-5pt}
Modern DL models are mostly trained using NVIDIA GPUs, because the primary computational steps (e.g., matrix-matrix multiplications) in DL match the SIMD operation that could be efficiently performed by GPUs.
In practice, DL practitioners often use single-node software frameworks, such as Caffe~\cite{Jia:2014:MM} and Torch~\cite{Collobert:2011:NIPSW}, which mathematically derive the correct training algorithm and execute it on GPU by calling GPU-based acceleration libraries, such as CUBLAS and cuDNN. It is thus straightforward to parallelize these programs across distributed GPUs using either PS or SFB, by moving the computation from CPU to GPU, and performing memory copy operations (between DRAM and GPUs) or communication (among multiple nodes) whenever needed. However, we argue below and show empirically in Section~\ref{sec:evaluation} that these usually lead to suboptimal performance.

The inefficiency is mainly caused by parameter synchronization via the network. Compared to CPUs, GPUs are an order of magnitude more efficient in matrix computations; the production of gradients on GPUs is much faster than they can be naively synchronized over the network. As a result, the training computations are usually bottlenecked by communications.
For example, when training AlexNet~\cite{Krizhevsky:2012:NIPS} (61.5M parameters) on Titan X with a standard batch size $256$, 240 million gradients will be generated per second on each GPU (0.25s/batch). If we parallelize the training on 8 nodes using a PS, with every node also holding $1/8$ of parameters as a PS shard; then, every node needs to transfer $240$M$ \times 7/8 \times 4 = 840$M float parameters in one second to make sure the next iteration of computation not being blocked. 
Apparently, the demanded throughput ($>$26Gbps) exceeds the bandwidth that commodity Ethernet (i.e., 1GbE and 10GbE Ethernet) provides; the GPUs distributed across clusters cannot be fully utilized. Practically, it is usually difficult to partition the parameters completely equally, which will result in more severe bandwidth demands, or bursty communication traffic on several server nodes (as we will show in Section~\ref{sec:evaluation:comparison_to_others}), which prevents the trivial realization of efficient DL on distributed GPUs
\footnote{Frequent memory copy operations between DRAM and GPU memory can also cause extra overheads, which is minor compared to the network communication according to our empirical results. However, our strategies in this paper can also alleviate this overhead.}.
We next describe our strategies and system design to overcome the aforementioned obstacles.

\vspace{-15pt}
\section{Poseidon Design}
\label{sec:design}
\vspace{-8pt}
In this section, we first analyze the DL program in both a single-node and distributed environment by decomposing the program into a sequence of operations. Based on it, we introduce two strategies to address the issues. 

\begin{figure}[t]
\centering
\includegraphics[width=0.7\columnwidth]{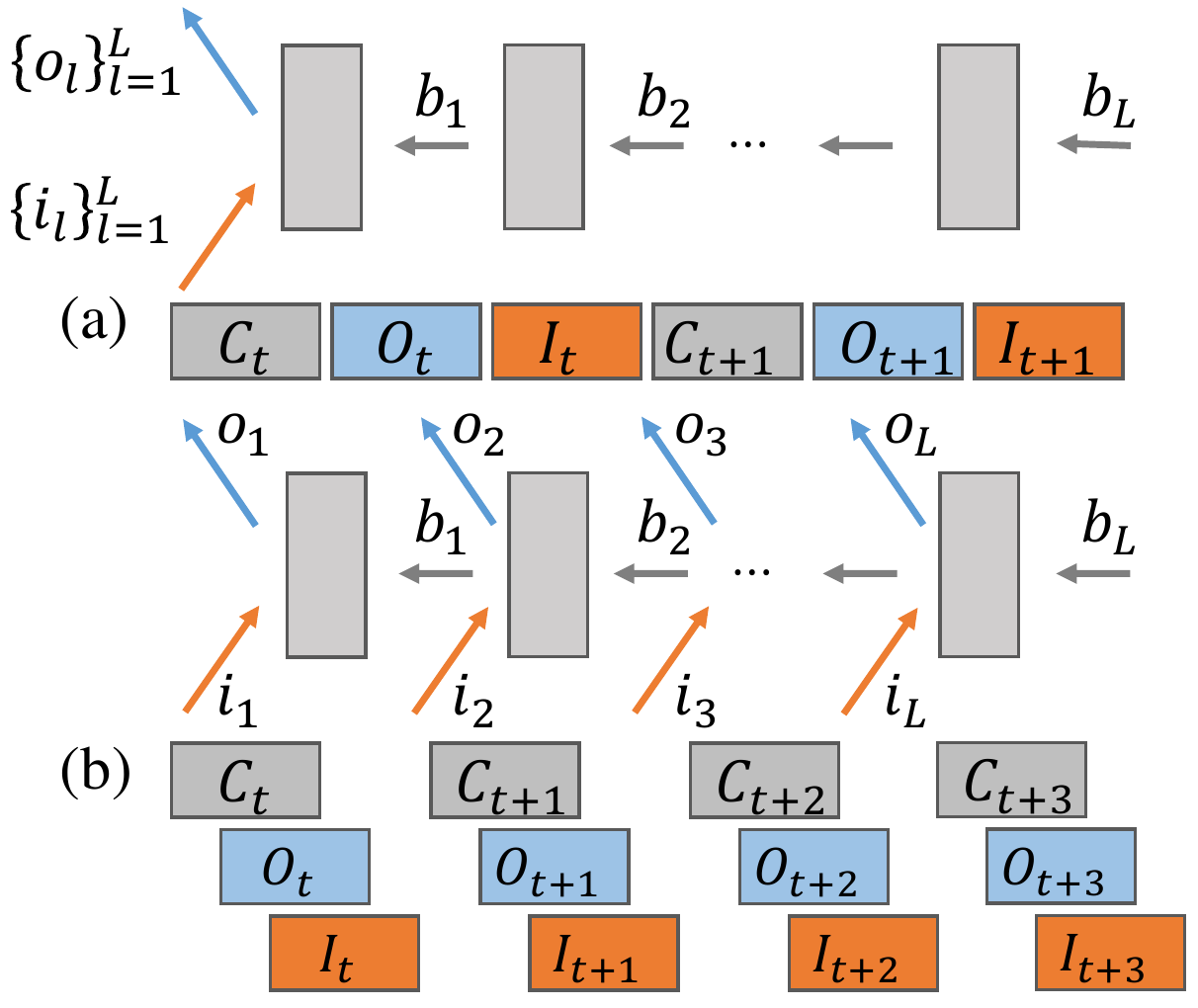}
\vspace{-10pt}
\caption{(a) Traditional backpropagation and (b) wait-free backpropagation on distributed environment.}
\label{fig:structure}
\vspace{-18pt}
\end{figure}

\noindent \textbf{The Structure of DL Programs.}
At the core of the DL program is the BP algorithm that performs forward-backward pass through the network repeatedly. If we define a forward and a backward pass through the $l$th layer of a network as $f_t^l$ and $b_t^l$, respectively, then a \textbf{C}omputation step at iteration $t$ is notated as $C_t = [f_t^1, \cdots, f_t^L, b_t^{L}, \cdots, b_t^{1}]$, as illustrated in Fig.~\ref{fig:structure}(a). 
When executing on distributed GPUs, inter-machine communications are required after each $C$ step to guarantee the synchronized replication of model parameters. We similarly define the \textbf{S}ynchronization step $S_t$ as the process that a worker sends out locally generated  updates and then receives updated parameters from remote workers at iteration $t$.
Therefore, a naive parallelization of DL training over distributed GPUs using either PS or SFB can be expressed as alternating $C_t$ and $S_t$ defined above. We note that DL training is highly sequential; the communication and computation perform sequentially, waiting each other to finish (Fig.~\ref{fig:structure}a).

Fortunately, we also note that as every layer of a NN contains an independent set of parameters, $S_t$ can be decoupled as $S_t = (s_t^1, \cdots, s_t^L)$, by defining $s_t^l$ as the synchronization of parameters of layer $l$. If we further decompose $s_t^l = [o_t^l, i_t^l]$ as first sending out local updates of layer $l$  ($o_t^l$) and reads in the updated parameters remotely ($i_t^l$), we can rewrite a training iteration as: $[C_t, S_t] = [f_t^1, \cdots, f_t^L, b_t^{L}, \cdots, b_t^{1}, o_t^L, \cdots, o_t^1, i_t^L, \cdots, i_t^1]$.
The sequential nature of the BP algorithm presents us an opportunity to overlap the computations and communications. Our first strategy, \emph{wait-free backpropagation}, overlaps $C_t$ and $S_t$ by partially rescheduling those $b_t$ and $s_t$ that are independent. Our second strategy, \emph{hybrid communication}, utilizes the independency among $s_t$, and tries to reduce the communication overheads by specializing different communication methods for different $s_t$.

\vspace{-15pt}
\subsection{Wait-free Backpropagation}
\vspace{-5pt}
The wait-free backpropagation (WFBP) is designed to overlap communication overheads with the computation based on two key independencies in the program: (1) the send-out operation $o_t^l$ is independent of backward operations $b_t^i (i < l)$, so they could be executed concurrently without blocking each other; (2) the read-in operation $i_t^l$ could update the layer parameters as long as $b_t^l$ was finished, without blocking the subsequent backward operations $b_t^i (i < l)$.
Therefore, we can enforce each layer $l$ to start its communication once its gradients are generated after $b_t^l$, so that the time spent on operation $s_t^l$ could be overlapped with those of $b_t^i (i < l)$, as shown in Fig.~\ref{fig:structure}b.

WFBP is most beneficial for training DL models that have their parameters concentrating at upper layers (FC layers) but computation concentrating at lower layers (CONV layers)\footnote{Most classification models will fall into this family if the number of classes to be classified is large.}, e.g., VGG~\cite{Simonyan:2015:ICLR} and AdamNet~\cite{Chilimbi:2014:OSDI,cui2016geeps}), because it overlaps the communication of top layers (90\% of communication time) with the computation of bottom layers (90\% of computation time)~\cite{zhang2015poseidon,cui2016geeps}.
Besides chain-like NNs, WFBP is generally applicable to other non-chain like structures (e.g., tree-like structures), as the parameter optimization for deep neural networks depends on adjacent layers (and not the whole network), there is always an opportunity for parameter optimization (i.e., computation) and communication from different layers to be performed concurrently. 

Some DL frameworks, such as TensorFlow, represent the data dependencies of DL programs using graphs, therefore implicitly enable auto-parallelization. However, they fail on exploring the potential opportunities of parallelization between iterations. For example, TensorFlow needs to fetch the updated parameters from the remote storage at the beginning of each iteration, while it is possible to overlap this communication procedure with the computation procedure of the previous iteration. In comparison, WFBP enforces this overlapping by explicitly pipelining compute, send and receive procedures. 
We describe our implementation of WFBP in Section~\ref{sec:implementation} and empirically show its effectiveness in Section~\ref{sec:evaluation:scalability}.

\vspace{-15pt}
\subsection{Hybrid Communication}
\vspace{-5pt}
\label{sec:hybComm}
While WFBP overlaps communication and computation, it does not reduce the communication overhead. In situations where the network bandwidth is limited (e.g., commodity Ethernet or the Ethernet is shared with other communication-heavy applications), the communication would still be unacceptably slow. To address the issue, we introduce a \emph{hybrid communication} (HybComm) strategy that combines the best of PS and SFB by being aware of both the mathematical property of DL models and the structure of computing clusters.
Our idea comes from two observations: first, as presented in Section~\ref{sec:design}, the synchronization operations $\{S_t^l\}_{l=1}^L$ are independent of each other, meaning that we can use different communication methods for different $S_t^l$ by specializing $o_t^l$ and $i_t^l$ according to the two methods described in Figure~\ref{fig:pssfb}; second, a NN structure is usually predefined and fixed throughout the training -- by measuring the number of parameters needed to transferred, we are able to estimate the communication overhead, so that we can always choose the optimal method even before the communication happens. 

Consider training VGG19 network~\cite{Simonyan:2015:ICLR}, the overheads of $S_t^l$ could be estimated as follows (Table~\ref{tb:cost}): assume the batch size $K = 32$, the number of workers and server nodes $P_1 = P_2 = 8$ (assume parameters are equally partitioned over all server shards), respectively. On one hand, if $l$ is an FC layer (with shape $4096 \times 4096, M = N = 4096$), synchronizing its parameters via PS will transfer $2MN \approx 34$ million parameters for a worker node, $2P_1MN / P_2 \approx 34$ million for a server node, and $2MN(P_1 + P_2 - 2)/P_2 \approx 58.7$ million for a node that is both a server and a worker, compared to $2K(M + N)(P_1-1) \approx 3.7$ million for a single node using SFB. On the other hand, if $l$ is a CONV layer, the updates are indecomposable and sparse, so we can directly resort to PS. Therefore, the synchronization overheads depend not only on the model (type, shape, size of the layer), but also the size of the clusters. The optimal solution usually changes with $M, N, K, P_1, P_2$. HybComm takes into account these factors and allows to dynamically adjust the communication method for different parts of a model -- it always chooses the best method from available ones whenever it results in fewer communication overheads.
\begin{table}[bt]
\footnotesize
\centering
	\begin{tabular}{|K{0.7cm}|K{1.6cm}|K{1.6cm}|K{2cm}|}
	\hline
	Method & Server & Worker & Server \& Worker \\ \hline \hline
	\textbf{PS} & $2P_1MN/P_2$ & $2MN$ & $2MN(P_1 + P_2 - 2)/P_2$  \\ \hline
  	\textbf{SFB} & \textbf{N/A} & $2 K (P_1 - 1)(M +N)$  & \textbf{N/A}  \\ \hline
  	\textbf{Adam} (max) & $P_1MN + P_1K(M+N) $  & $K(M+N)+MN$ &  $(P_1 - 1) (MN + KM + KN)$  \\ \hline
    \end{tabular}
    \vspace{-5pt}
\caption{\small Estimated communication cost of PS, SFB and Adam for synchrnizing the parameters of a $M \times N$ FC layer on a cluster with $P_1$ workers and $P_2$ servers, when batchsize is $K$.}
\vspace{-20pt}
\label{tb:cost}
\end{table}

Microsoft Adam~\cite{Chilimbi:2014:OSDI} employs a different communication strategy from those in Figure~\ref{fig:pssfb}. Instead of broadcasting SFs across workers, they first send SFs to a parameter server shard, then pull back the whole updated parameter matrices. This seems to reduce the total number of parameters needed to be communicated, but usually leads to load imbalance; the server node that holds the corresponding parameter shard overloads because it has to broadcast the parameter matrices to all workers ($P_1 MN + P_1 K (M+N)$ messages need to be broadcasted), which easily causes communication bottleneck (Section~\ref{sec:evaluation:comparison_to_others}).
It is noticeable that reconstructing gradients from SFs may cause extra computation cost, which however is often negligible compared to communication. We describe our implementation of HybComm in the next section, and assess its effectiveness in Section \ref{sec:evaluation}.

\vspace{-15pt}
\section{Implementation}
\label{sec:implementation}
\vspace{-10pt}
This section first elaborates Poseidon's system architecture and APIs, and then describes how to modify a framework using Poseidon to enable distributed execution.

\begin{table*}[h]
\footnotesize
\centering
	\begin{tabular}{|L{1.6cm}|L{1.2cm}|L{3.8cm}|L{8.2cm}|}
	\hline
	\textbf{Method} & \textbf{Owner} & \textbf{Arguments} & \textbf{Description} \\ \hline \hline
	\texttt{BestScheme} &  Coordinator & A layer name or index & Get the best communication scheme of a layer \\ \hline
      \texttt{Query} & Coordinator & A list of property names & Query information from coordinators' information book \\ \hline
  	\texttt{Send} & Syncer & None & Send out the parameter updates of the corresponding layer \\ \hline
  	\texttt{Receive} & Syncer & None & Receive parameter updates from either parameter server or peer workers \\ \hline
    \texttt{Move} & Syncer & A GPU stream and an indicator of move direction & Move contents between GPU and CPU, do transformations and application of updates if needed \\ \hline
    \texttt{Send} & KV store & updated parameters &Send out the updated parameters \\ \hline
  	\texttt{Receive} & KV store & parameter buffer of KV stores & Receive gradient updates from workers  \\ \hline
    \end{tabular}
    \vspace{-8pt}
\caption{\small Poseidon APIs for parameter synchronization.}
\vspace{-18pt}
\label{tb:api}
\end{table*}

\begin{algorithm}[h]
\begin{algorithmic}[1]
\small
\Function{BestScheme}{$l$}
\State $layer\_property$ = Query($l$.name)
\State $P_1, P_2, K$ = Query(`n\_worker', `n\_server', `batchsize')
\If{$layer\_property$.type == `FC'}
\State     $M$ = $layer\_property$.width 
\State     $N$ = $layer\_property$.height
\If{ $2K(P_1 - 1)(M+N) \le \frac{2MN(P_1 +  P_2 - 2)} { P_2}$ }
\State   return `SFB'
\EndIf
\EndIf
\State return `PS'
\EndFunction
\end{algorithmic}
\caption{Get the best comm method of layer $l$}
\label{algo:best_method}
\end{algorithm}

\begin{figure}[h]
\centering
\vspace{-12pt}
\includegraphics[width=0.9\columnwidth]{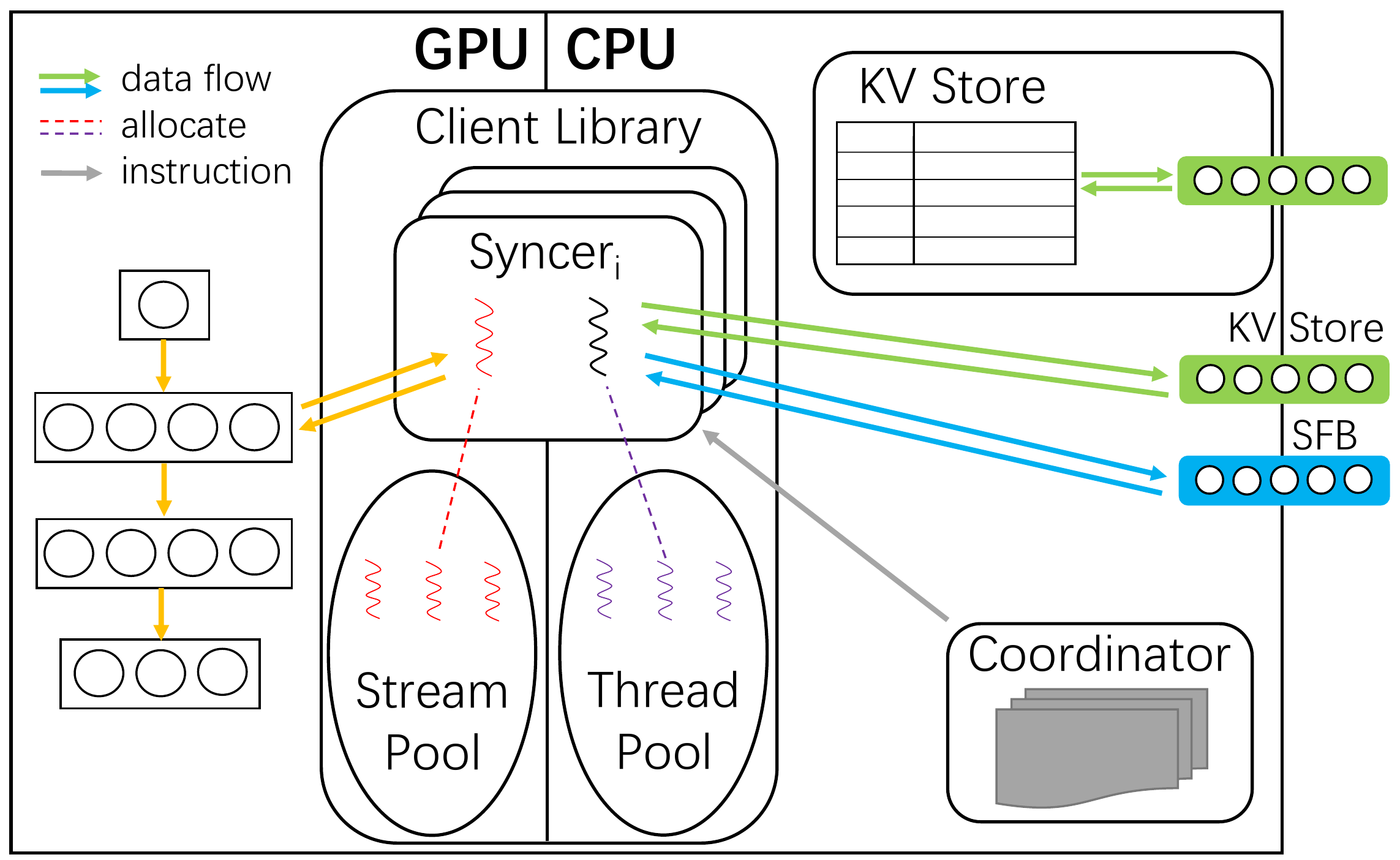}
\vspace{-10pt}
\caption{\small An overview of the architecture of Poseidon.}
\vspace{-22pt}
\label{fig:architecture}
\end{figure}

\vspace{-12pt}
\subsection{System Implementation and APIs}
\vspace{-5pt}
Figure \ref{fig:architecture} illustrates the architecture of Poseidon: a C++ communication library that manages parameter communication for DL programs running on distributed GPUs.
It has three main components: coordinator, that maintains the model and the cluster configuration; KV store, a shared memory key-value store that provides support for parameter server based communication; client library, which is plugged into DL programs to handle parameter communication. Their APIs are listed in Table~\ref{tb:api}.

\noindent \textbf{Coordinator.}
To setup distributed training, the client program (e.g., Caffe) first instantiates Poseidon by creating a coordinator within its process.
Coordinators will first collect necessary information, including the cluster information (e.g., the number of workers and server nodes, their IP addresses) and the model architecture (e.g., the number of layers, layer types, number of neurons and how they are connected, etc.). 
With the information, the coordinator will initialize the KV stores and the client library with two steps: (1) allocate proper communication ports for each PS shard and peer worker; (2) determine what parameters should be transmitted via the KV store and what by SFB, and hash the parameters equally to each KV store if necessary, and save the mapping in the information book, 
which, throughout the whole training, is maintained and synchronized across nodes, and could be accessed elsewhere through coordinator's \texttt{Query} API.
Besides, the coordinator provides another API \texttt{BestScheme} that takes in a layer and returns the optimal communication scheme for it according to the strategy described in Section~\ref{sec:hybComm} (Algorithm~\ref{algo:best_method}).

\noindent \textbf{KV Store.}
The KV store is implemented based on a bulk synchronous parameter server~\cite{Wei:2015:SoCC,cui2016geeps}, and instantiated by coordinators on a list of user-specified ``server'' machines. Each instance of the KV store holds one shard of the globally shared model parameters in the form of a set of KV pairs, of which each KV pair is stored on a chunk of DRAM. Poseidon sets the size of a KV pair to a fixed small size (e.g., 2MB), so as to partition and distribute model parameters to server nodes as equally as possible, reducing the risk of Ethernet bottleneck.
Each KV store instance manages a parameter buffer on RAM, and provides PS-like APIs, such as \texttt{Receive} and \texttt{Send}, for receiving and applying updates from client libraries, or sending out parameters. It will regularly checkpoint current parameter states for fault tolerance.

\noindent \textbf{Client Library.}
Poseidon coordinates with DL programs via its client library.
Particularly, users plug the client library into their training program, and the client library will create a \emph{syncer} for each NN layer during network assembling (so that each layer one-to-one maps to one syncer), accounting for its parameter synchronization.
Each sycner is then initialized, for example, setting up connections to its corresponding PS shards or (remote) peer syncers according to the coordinator's information book, and allocating a small memory buffer for receiving remote parameter matrices or SFs, etc.

The client library manages a CPU thread pool and a GPU stream pool on the worker machine, which can be allocated by the syncer APIs when there is a syncer job created.
The syncer has three main APIs, \texttt{Send}, \texttt{Receive} and \texttt{Move}, to be used in client programs.
The \texttt{Move} API takes care of the memory movement between RAM and GPU memory, and performs necessary computation, e.g., the transformation between SFs and gradients, and the application of updates. It is multi-threaded using the CUDA asynchronous APIs, and will trigger an allocation from the client library's thread/stream pools when a syncer job starts (see L14 of Algorithm~\ref{algo:plugin}).
The \texttt{Send} and \texttt{Receive} are communication APIs that synchronize layer parameters across different model replicas. The \texttt{Send} API is nonblocking; it sends out parameter updates during backpropagation once they are generated, following the protocol returned by coordinator's \texttt{BestScheme} API.
The \texttt{Receive} API will be called once \texttt{Send} is finished. It requests either fresh parameter matrices from the KV stores or SFs from its peer syncers, and will block its current thread until it receives all of what it requested. The received messages are put into the syncer's memory buffer for the \texttt{Move} API to fetch.

\noindent \textbf{Managing Consistency.}
Poseidon implements the bulk synchronous consistency (BSP) model as follows.
The client library maintains a binary vector $C$ with length the number of syncers and values reset to zeros at the start of each iteration. 
A syncer will set its corresponding entry in $C$ as 1 when its job finishes, and the client starts next iteration when all entries are 1.
While, the KV store maintains a zero-initialized count value for each KV pair at the start of each iteration. Every time when there is an update being applied on a KV pair, its count value is increased by 1. 
The KV pair will be broadcasted via its \texttt{Send} API when its count equals to the number of workers.
Poseidon handles stragglers by simply dropping them. Although asynchronous models can alleviate the straggler problem in distributed ML~\cite{Ho:2013:NIPS}, Poseidon focuses on synchronous parallel training, because synchronous execution yields the fastest per-iteration improvement in accuracy for distributed DL (as measured by wall clock time) on GPUs~\cite{cui2016geeps,chen2016revisiting} (see Section~\ref{sec:evaluation:scalability}).
\vspace{-8pt}
\begin{algorithm}[]
\begin{algorithmic}[1]
\small
\Function{TRAIN}{$net$}
\For{$iter = 1 \rightarrow T$}
\State	$sync\_count=0$
\State	$net$.Forward()
    \For{$l = L \rightarrow 1$}
\State   	$net$.BackwardThrough($l$)
\State		$thread\_pool$.Schedule(sync($l$))
\EndFor
\State wait\_until($sync\_count == net.num\_layers$)
\EndFor
\EndFunction
\Function{sync}{$l$}
\State $stream$ = $stream\_pool$.Allocate()
\State $syncers[l]$.Move($stream$, GPU2CPU)
\State $syncers[l]$.$method$ = $coordinator$.BestScheme($l$)
\State $syncers[l]$.Send()
\State $syncers[l]$.Receive()
\State $syncers[l]$.Move($stream$, CPU2GPU)
\State $sync\_count$++
\EndFunction
\end{algorithmic}
\caption{Parallelize a DL library using Poseidon}
\label{algo:plugin}
\end{algorithm}

\vspace{-22pt}
\subsection{Integrate Poseidon with DL Libraries}
\vspace{-5pt}
Poseidon could be plugged into most existing DL frameworks to enable efficient distributed execution. Algorithm~\ref{algo:plugin} provides an example. Specifically, one needs to first include Poseidon's client library into the framework, then figure out where the backpropagation proceeds (L6), and insert Poseidon's syner APIs in between gradient generation and application (L7). We demonstrate in Section~\ref{sec:evaluation:scalability} that with slight modifications (150 and 250 LoC for Caffe and TensorFlow), both Poseidon-enable Caffe and TensorFlow deliver linear scalings up to 32 GPU machines. Poseidon respects the programming interfaces by the native DL library and stores necessary arguments for distributed execution as environment variables to allow zero changes on the DL application programs.

\vspace{-15pt}
\section{Evaluation}
\label{sec:evaluation}
\vspace{-10pt}
In this section, we evaluate Poseidon's performance on scaling up DL with distributed GPUs. We focus on the image classification task where DL is most successfully applied. Our evaluation reveals the following results: (1) Poseidon has little overhead when plugged into existing frameworks; it achieves near-linear speedups across different NNs and frameworks, on up to 32 Titan X-equipped machines. (2) Poseidon's system design effectively improves GPU and bandwidth utilization. (3) Poseidon's communication strategy HybComm effectively alleviates the communication bottleneck, thus achieves better speedups under limited bandwidth; 
Moreover, Poseidon compares favorably to other communication-reduction methods, such as the SF strategy in Adam~\cite{Chilimbi:2014:OSDI}, and the 1-bit quantization in CNTK~\cite{yu2014introduction}.

\noindent \textbf{Cluster Configuration.}
We conduct our experiments on a GPU cluster with each node equipped with a NVIDIA GeForce TITAN X GPU card, an Intel 16-core CPU and 64GB RAM, interconnected via a 40-Gigabit Ethernet switch. All cluster nodes have shared access to a NFS and read data through the Ethernet interface. We run our system on UBUNTU 16.04, with NVIDIA driver version 361.62, CUDA 8.0 and cuDNN v5.

\noindent \textbf{Computation Engines.}
We deploy Poseidon on two DL frameworks, Caffe~\cite{Jia:2014:MM} and TensorFlow~\cite{abadi2016tensorflow}.
For Caffe, we use the official version at 2016/06/30 as the single node baseline, and modify it using Poseidon's client library API for distributed execution. For TensorFlow, we use its open source version r0.10, and parallelize its single-node version with Poseidon's client library, and compare to its original distributed version. \footnote{Note that as the distributed engine of TensorFlow is highly optimized (e.g., auto-parallelization of graphs~\cite{abadi2016tensorflow}). Poseidon avoids leveraging any build-in optimization of distributed TensorFlow by parallelizing its single-node version instead.}

\noindent \textbf{Dataset and Models.}
Our experiments use three well-known image classification datasets. 
(1) CIFAR-10~\cite{Krizhevsky:2009:cifar},  which contains $32 \times 32$ colored images of $10$ classes, with 
50K images for training and 10K for testing;
(2) ILSVRC12~\cite{Russakovsky:2015:IJCV}, a subset of ImageNet22K that has 1.28 million of training images and 50K validation images in 1,000 categories;
(3) ImageNet22K~\cite{Russakovsky:2015:IJCV}, the largest public dataset for image classification, including 14,197,087 labeled images from 21,841 categories.

We test Poseidon's scalability across different neural networks:
(1) CIFAR-10 quick: a toy CNN from Caffe that converges at $73\%$ accuracy for classifying images in CIFAR-10 dataset;
(2) GoogLeNet~\cite{Szegedy:2014:going}: a 22-layer CNN with 5M parameters. 
(3) Inception-V3~\cite{szegedy2015rethinking}: the ImageNet winner, an improved version of GoogLeNet from TensorFlow;
(4) VGG19: A popular feature extraction network in the computer vision community~\cite{Simonyan:2015:ICLR} that has 16 CONV layers and 3 FC layers, in total 143M parameters;
(5) VGG19-22K: we modify the VGG19 network by replacing its 1000-way classifier with a 21841-way classifier, to classify images from the ImageNet22K dataset. The modified network has 229M parameters.
(6) ResNet-152: the ImageNet winner network with 152 layers.
We list their statistics and configurations in Table~\ref{tb:model_stats}.

\noindent \textbf{Metrics.} In this paper, we mainly focus on metrics that measure the system performance, such as speedups on throughput (number of images scanned per second). Our experiments focus on medium-scale distributed cluster with up to 32 machines, which distributed DL empirically benefits most from. Larger clusters require larger batch sizes, which hurt the convergence rate of each iteration~\cite{chen2015mxnet,cui2016geeps}. For completeness, we also report the statistical performance (time/epoch to converge) on ResNet-152. Poseidon uses synchronized replication which enables many models to converge in fewer steps~\cite{abadi2016tensorflow,cui2016geeps,chen2015mxnet,chen2016revisiting}.

\vspace{-10pt}
\subsection{Scalability}
\label{sec:evaluation:scalability}
\vspace{-5pt}
To demonstrate Poseidon's scalability, we train CNNs using Poseidon with different computational engines, and compare different systems in terms of their speedups on throughput. For Caffe engine, we train GoogLeNet
VGG19 and VGG19-22K networks; for TensorFlow engine, we train Inception-V3, VGG-19, VGG19-22K.

\begin{table}[]
\footnotesize
\centering
	\begin{tabular}{|c|c|c|K{1.5cm}|c|}
	\hline
	Model & \# Params & Dataset & Batchsize \\ \hline \hline
	\textbf{CIFAR-10 quick} & 145.6K & CIFAR10 & 100 \\ \hline
  	\textbf{GoogLeNet} & 5M &  ILSVRC12 & 128 \\ \hline
  	\textbf{Inception-V3} & 27M & ILSVRC12 &  32\\ \hline
  	\textbf{VGG19} & 143M & ILSVRC12 & 32 \\ \hline
  	\textbf{VGG19-22K} & 229M & ImageNet22K & 32\\ \hline
    \textbf{ResNet-152} & 60.2M & ILSVRC12 & 32 \\ \hline
    \end{tabular}
    \vspace{-8pt}
\caption{\small Neural networks for evaluation. Single-node batchsize is reported. The batchsize is chosen based on the standards reported in literature (usually the maximum batch size that can fill in the GPU memory).}
\vspace{-20pt}
\label{tb:model_stats}
\end{table}

\begin{figure*}[thbp]
\centering
        \begin{subfigure}[b]{0.31\textwidth}
                \includegraphics[width=\linewidth]{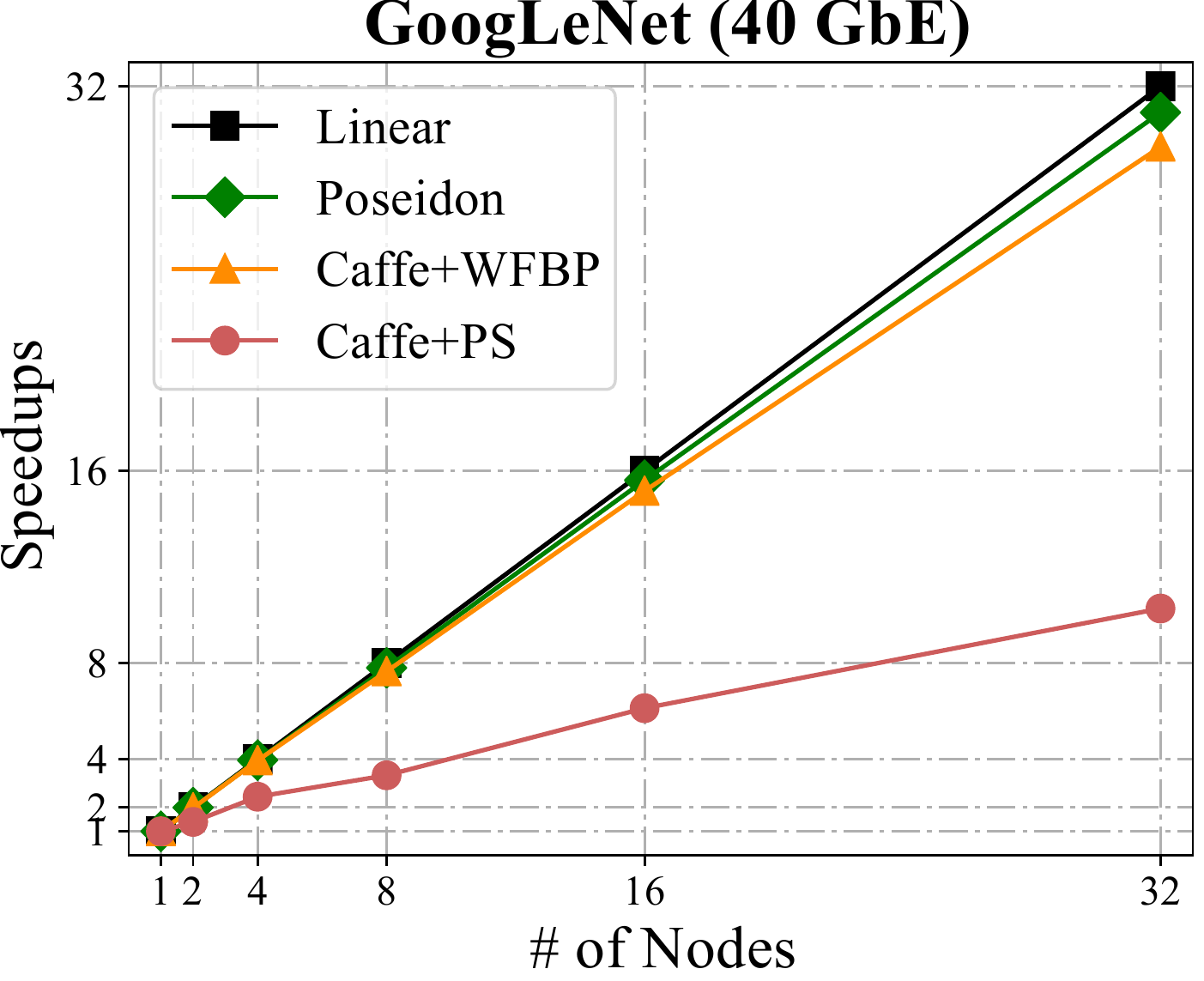}
        \end{subfigure}%
        \begin{subfigure}[b]{0.31\textwidth}
                \includegraphics[width=\linewidth]{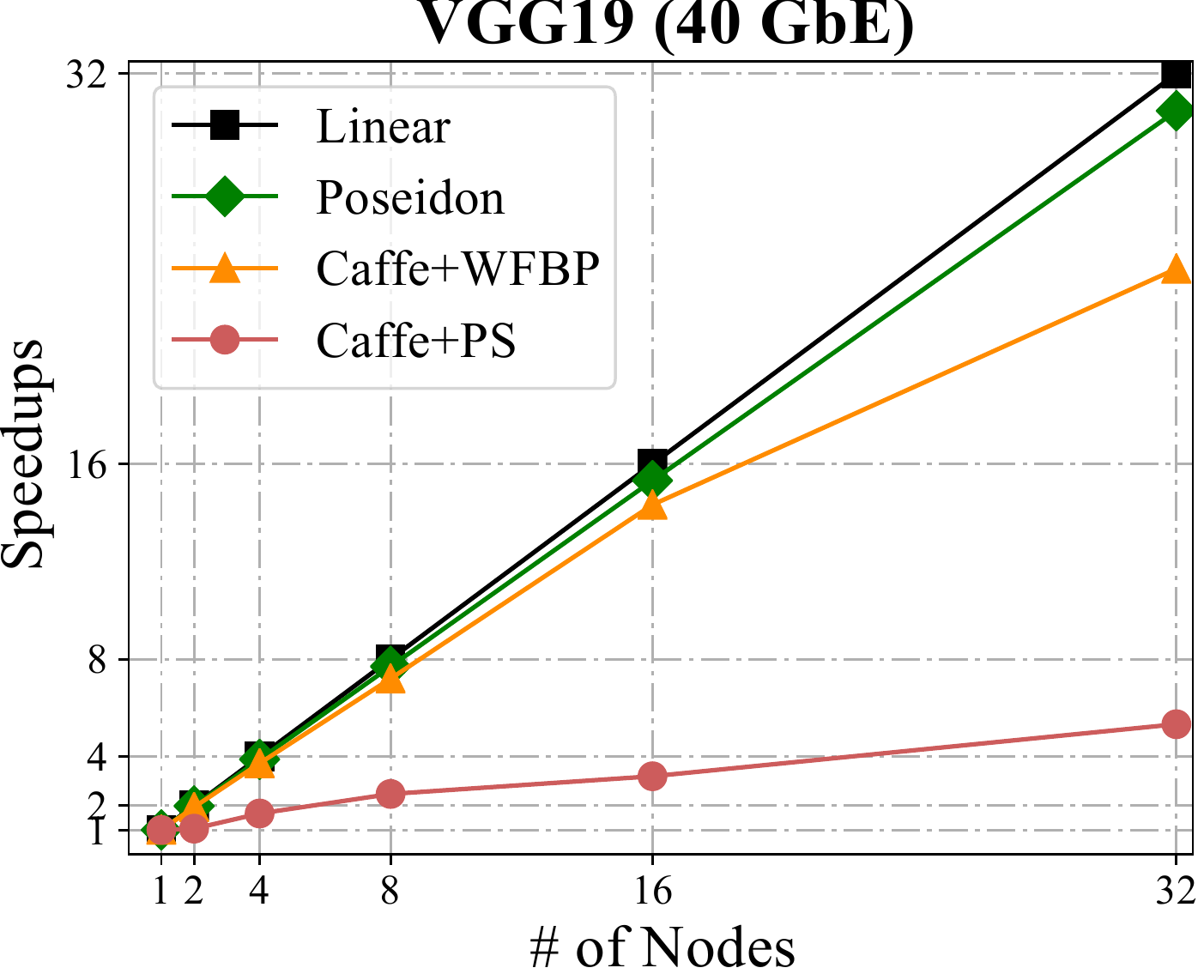}
        \end{subfigure}%
        \begin{subfigure}[b]{0.31\textwidth}
                \includegraphics[width=\linewidth]{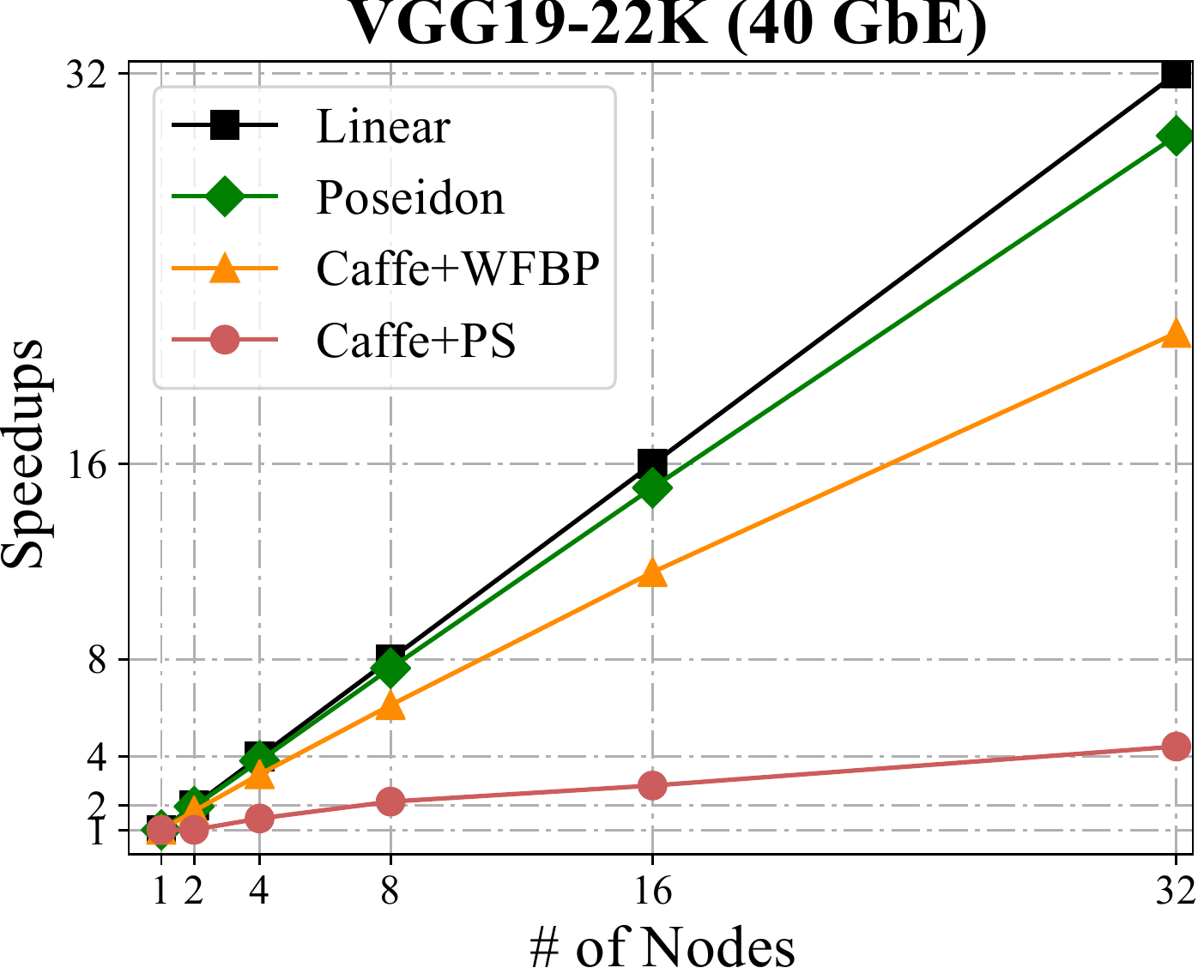}
        \end{subfigure}%
        \vspace{-10pt}
        \caption{Throughput scaling when training GoogLeNet, VGG19 and VGG19-22K using Poseidon-parallelized Caffe and 40GbE bandwidth. Single-node Caffe is set as baseline (i.e., speedup = 1).}\label{fig:scalability_caffe}
\end{figure*}

\noindent \textbf{Caffe Engine.}
Figure~\ref{fig:scalability_caffe} shows the throughput vs. number of workers when training the three networks using Caffe engine, given 40GbE Ethernet bandwidth available. We compare the following systems:
(1) \emph{Caffe}: unmodified Caffe that executes on a single GPU;
(2) \emph{Caffe+PS}: we parallelize Caffe using a vanilla PS, i.e., the parameter synchronization happens sequentially after the backpropagation in each iteration;
(3) \emph{Caffe+WFBP}: Parallelized Caffe using Poseidon so the communication and computation are overlapped. However, we disable HybComm so that parameters are synchronized only via PS;
(4) \emph{Poseidon}: the full version of Poseidon-Caffe.

Poseidon shows little overheads when combined with Caffe; running on a single node with no communication involved, Poseidon-Caffe can process 257, 35.5 and 34.2 images per second when training GoogLeNet, VGG19 and VGG19-22K, respectively, as compared to the original Caffe, which can process 257, 35.5 and 34.6 images, and Caffe+PS, which can only process 213.3, 21.3 and 18.5 images per second, due to the overheads caused by memory copy operations between RAM and GPU, which have been overlapped by Poseidon with the computation.
In distributed environment, the rescheduling of computation and communication significantly improves the throughput: when training GoogLeNet and VGG19, incorporating WFBP achieves almost linear scalings up to 32 machines, and for the larger VGG19-22K network, Caffe+WFBP achieves 21.5x speedup on 32 machines. We conclude that rescheduling and multi-threading the communication and computation are key to the performance of distributed DL on GPUs, even when the bandwidth resource is abundant. Poseidon provides an effective implementation to overlap these operations for DL frameworks, to guarantee better GPU utilization.

When the available bandwidth is sufficient, Poseidon's HybComm strategy shows small improvement on training GoogLeNet and VGG19. However, when training VGG19-22K which has three FC layers that occupy 91\% of model parameters, it improves over Caffe-WFBP from 21.5x to 29.5x on 32 nodes. 

\begin{figure*}[tbh]
\vspace{-10pt}
\centering
        \begin{subfigure}[b]{0.31\textwidth}
                \includegraphics[width=\linewidth]{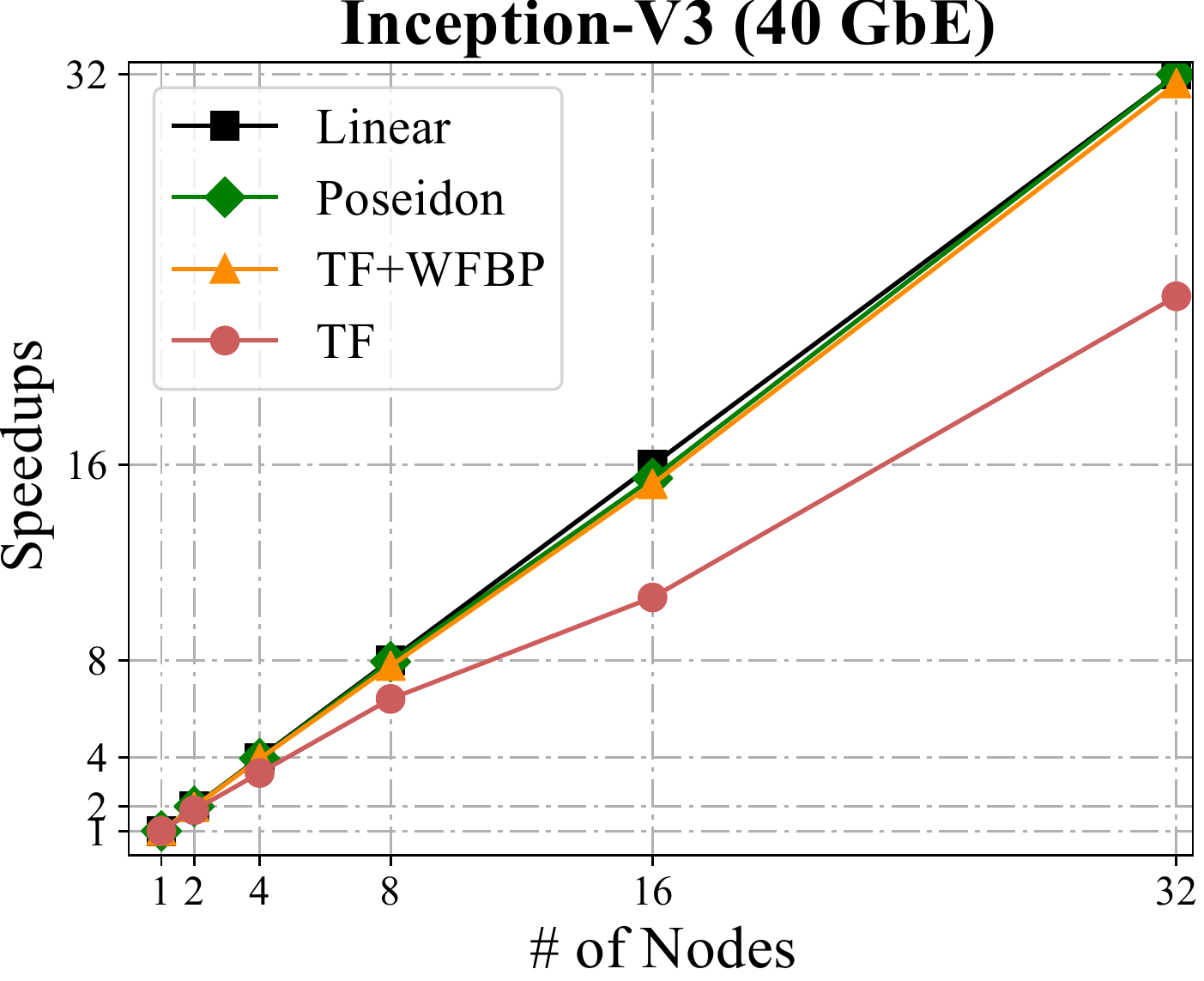}
        \end{subfigure}%
        \begin{subfigure}[b]{0.31\textwidth}
                \includegraphics[width=\linewidth]{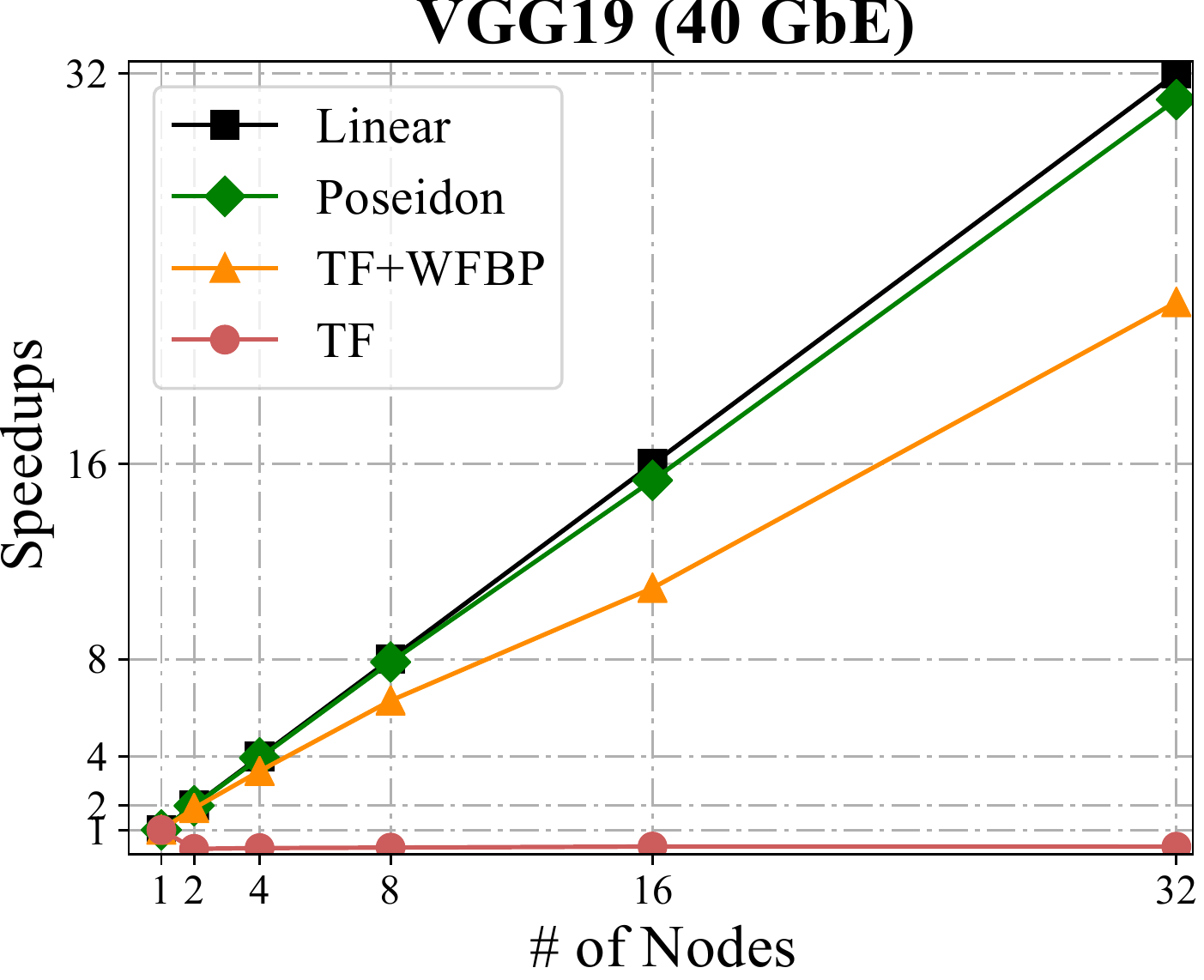}
        \end{subfigure}%
        \begin{subfigure}[b]{0.31\textwidth}
                \includegraphics[width=\linewidth]{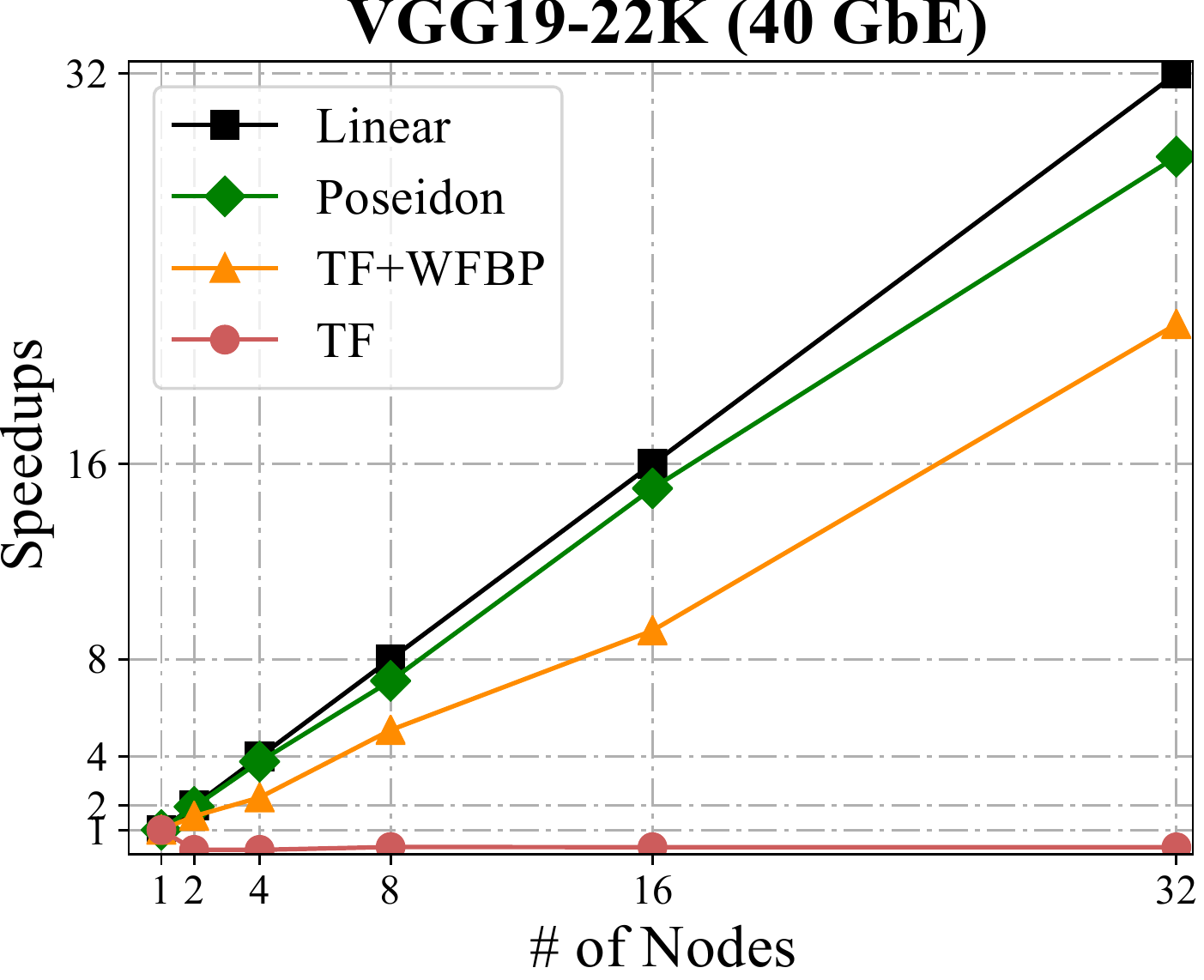}
        \end{subfigure}%
        \vspace{-10pt}
        \caption{\small Throughput scaling when training Inception-V3, VGG19 and VGG19-22K using Poseidon-parallelized TensorFlow and 40GbE bandwidth. Single-node TensorFlow is set as baseline (i.e., speedup = 1). }\label{fig:scalability_tf}
\vspace{-18pt}
\end{figure*}

\begin{figure}[tbh]
\centering
\includegraphics[width=0.9\linewidth]{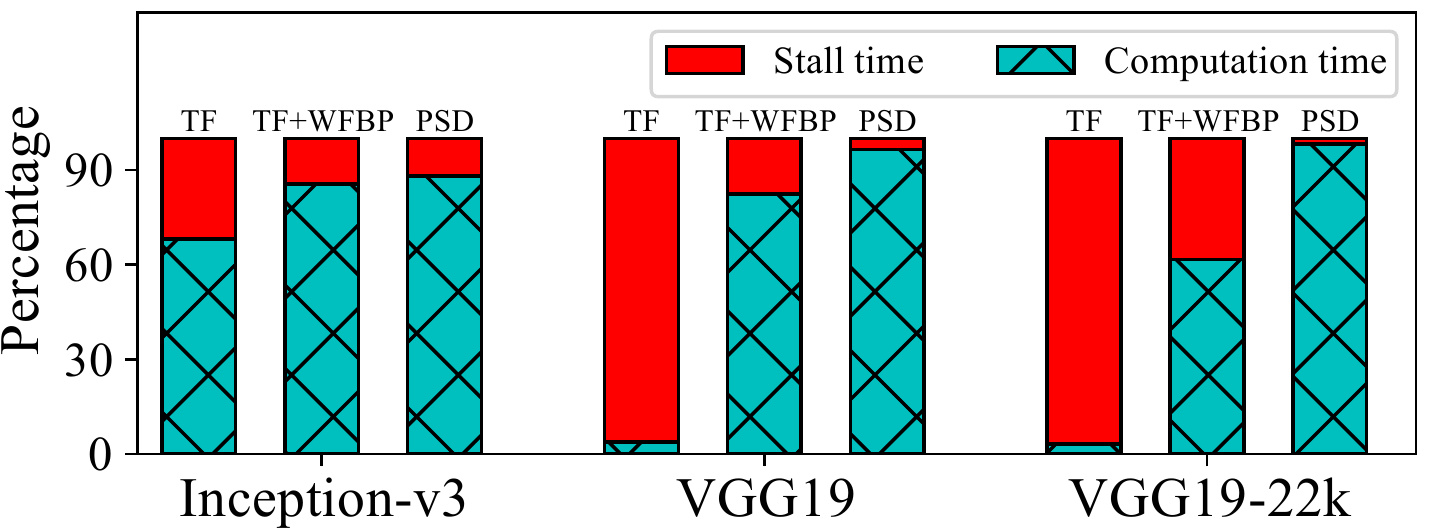}
\vspace{-10pt}
\caption{\small Breakdown of GPU computation and stall time when training the three networks on 8 nodes using different systems.}
\label{fig:compute_comm_tf}
\vspace{-22pt}
\end{figure}

\noindent \textbf{TensorFlow Engine.}
We also modify TensorFlow using Poseidon, and compare the following systems in terms of speedup on throughput:
(1) \emph{TF}: TensorFlow with its original distributed executions;
(2) \emph{TF+WFBP}: we modify TensorFlow using Poseidon's client library. Specifically, we change the \emph{assign} operator in TensorFlow, so that instead of being applied, the parameter updates will be synchronized via \emph{Poseidon's PS interface with WFBP};
(3) \emph{Poseidon}: the full version of Poseidon-parallelized TensorFlow with HybComm enabled.

We train Inception-V3, VGG19 and VGG19-22K models
and report the results in Figure~\ref{fig:scalability_tf}. Running on a single node, Poseidon processes 43.2, 38.2 and 34.5 images per second on training Inception-V3, VGG19 and VGG19-22K, while original TensorFlow processes 43.2, 38.5 and 34.8 images per second on these three models, respectively -- little overhead is introduced by our modification. In distributed execution, Poseidon achieves almost linear speedup on up to 32 machines. Distributed TensorFlow, however, demonstrates only 10x speedup on training Inception-V3 and even fails to scale on training the other two networks in our experiments. 
To investigate the problem of TensorFlow and explain how Poseidon improves upon it, we illustrates in Figure~\ref{fig:compute_comm_tf} the (averaged) ratio of busy and stall time of a GPU when training the three networks using different systems on 8 nodes. Observe that Poseidon keeps GPUs busy in most of the time, while TensorFlow wastes much time on waiting for parameter synchronization. The inefficiency of distributed TensorFlow stems from two sources. First, TensorFlow partitions model parameters in a coarse-grained granularity -- each tensor (instead of a KV pair) in the model is assigned to a PS shard. A big tensor (such as the parameter matrix in VGG19) is highly likely to create communication bottleneck on its located server node. Poseidon fixes this problem by partitioning parameters among server nodes in a finer-grained granularity using KV pairs, so that every node has evenly distributed communication load; as an evidence, TF-WFBP demonstrates higher computation-to-stall ratio in Figure~\ref{fig:compute_comm_tf}. 
Second, TensorFlow cannot reduce the communication overheads while Poseidon's HybComm effectively reduces the size of messages. As a result, Poseidon further improves upon TF-WFBP from 22x to 30x on 32 nodes. 

\noindent \textbf{Multi-GPU Settings.}
Poseidon's key strategies can be directly extended to support distributed multi-GPU environment with minor modifications. Specifically, when there are more than 1 GPU on a worker node, Poseidon will first collect the gradient updates following WFBP locally (either by full matrices or SFs) from multiple GPUs to a leader GPU using \emph{CudaMemcpy(DeviceToDevice)} API. If those updates are determined to be communicated via full matrices, Poseidon will aggregate them locally before sending out. 
Using Caffe engine on a single node, Poseidon achieves linear scalings on up to 4 Titan X GPUs when training all three networks, outperforming Caffe's multi-GPU version, which shows only 3x and 2x speedups when training GooLeNet and VGG19.
When running on AWS p2.8xlarge instances (8 GPUs each node), Poseidon reports 32x and 28x speedups when training GoogLeNet and VGG19 with 4 nodes (32 GPUs in total), confirming our statement that the overheads caused by memory movement between GPUs are usually negligible compared to network communication\footnote{The K80 GPUs on p2.8xlarge has less GFLOPS than Titan X used in our main experiments -- the communication burden is less severe.}.


\begin{figure*}[t]
\centering
        \begin{subfigure}[b]{0.31\textwidth}
                \includegraphics[width=\linewidth]{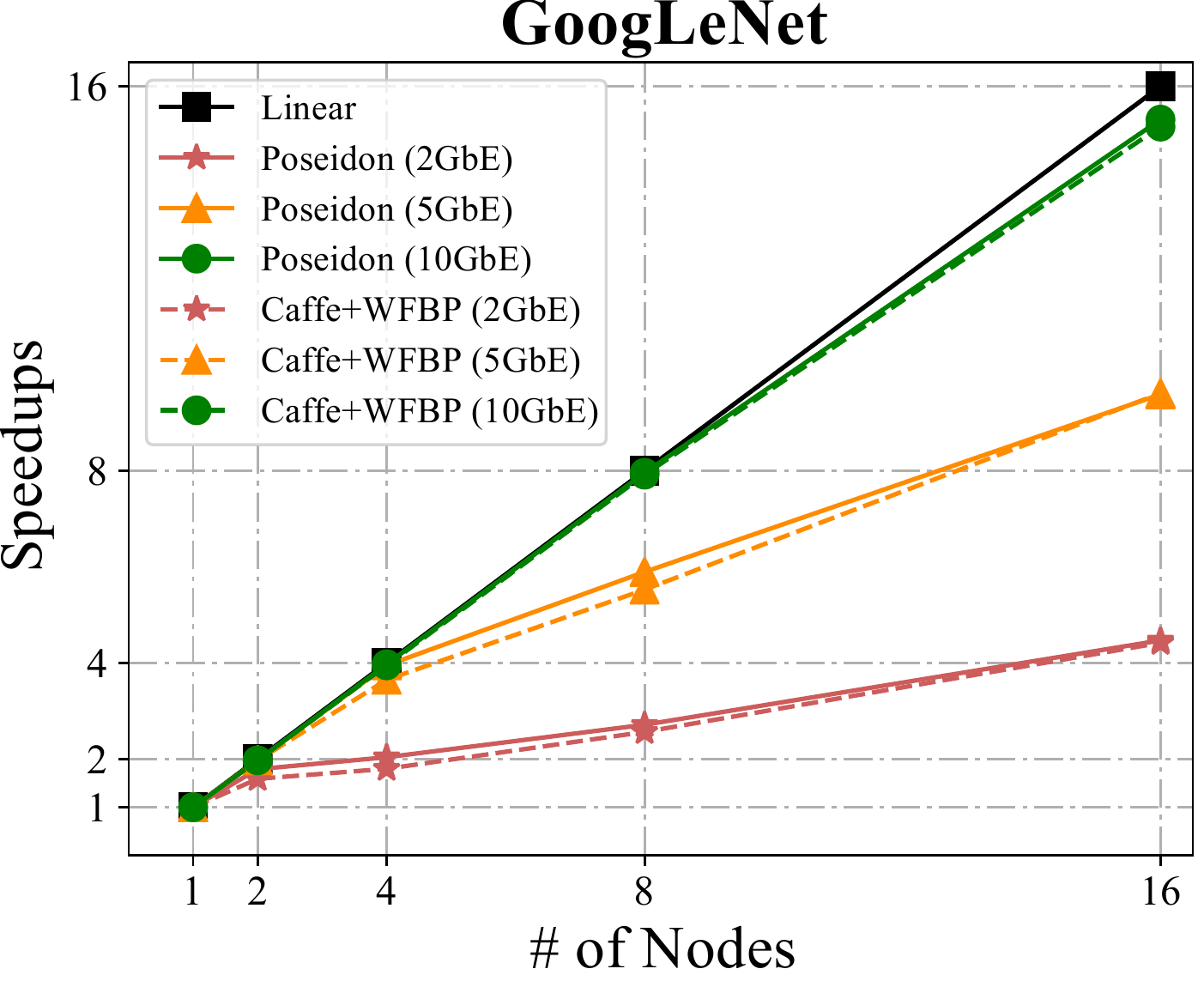}
        \end{subfigure}%
        \begin{subfigure}[b]{0.31\textwidth}
                \includegraphics[width=\linewidth]{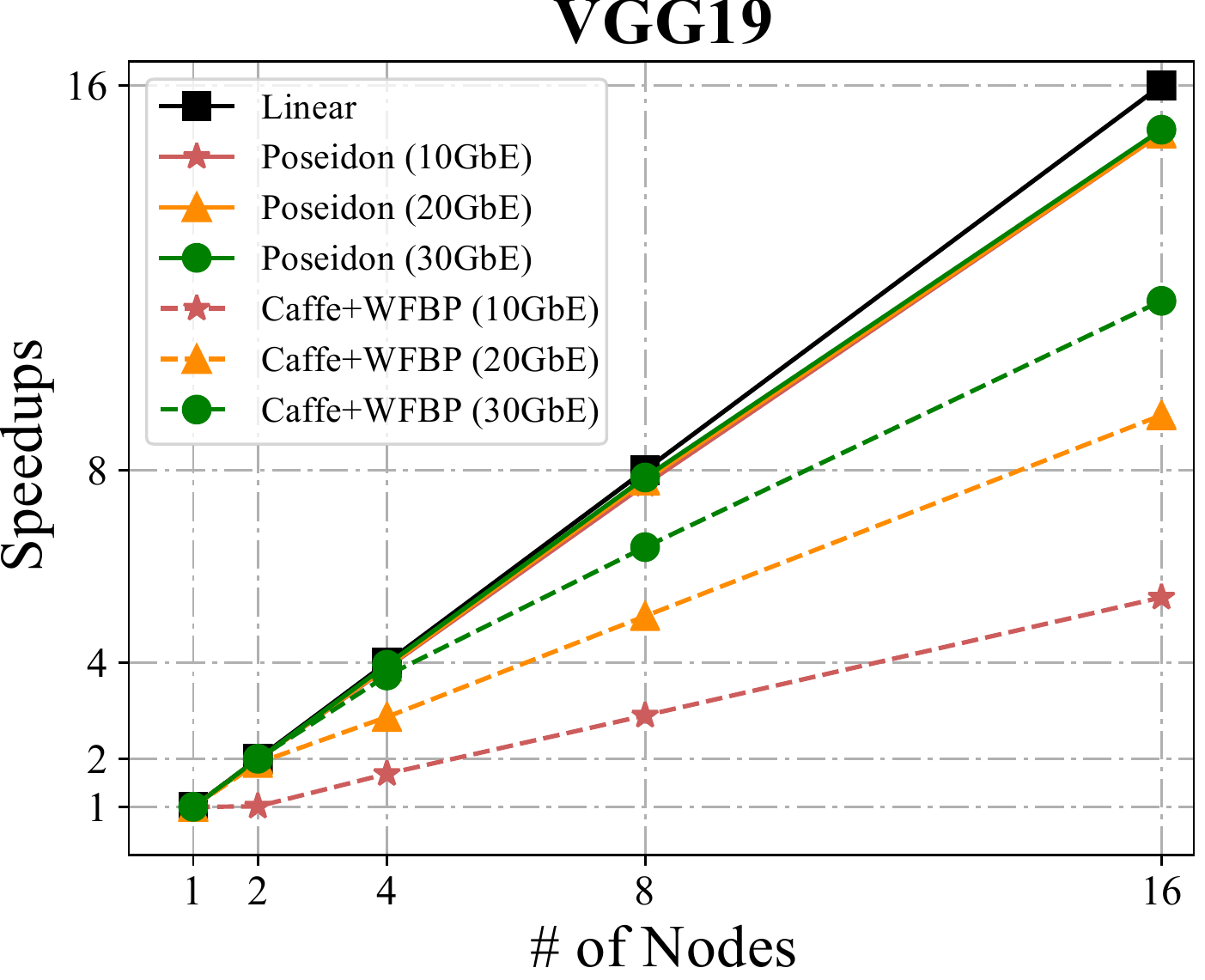}
        \end{subfigure}%
        \begin{subfigure}[b]{0.31\textwidth}
                \includegraphics[width=\linewidth]{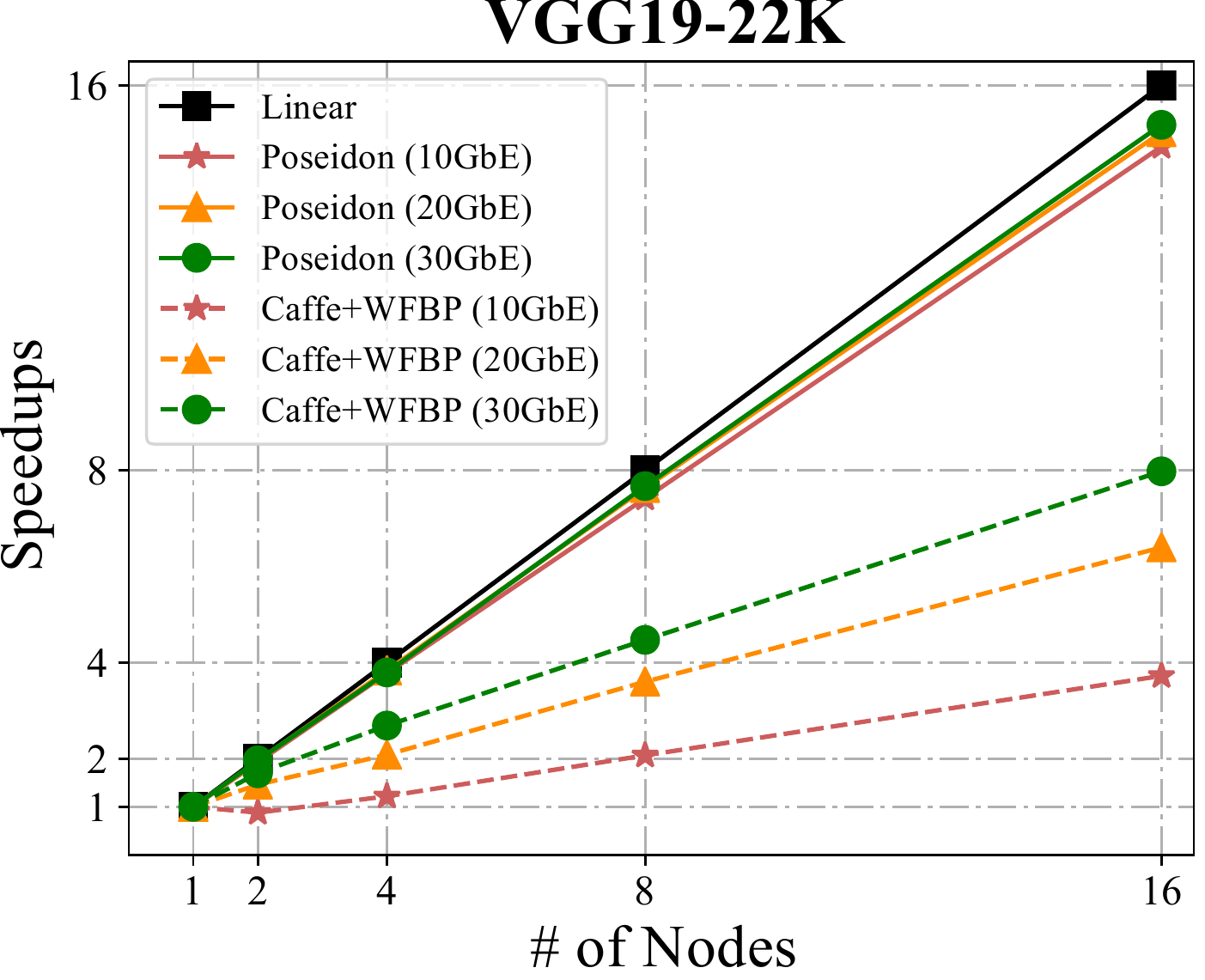}
        \end{subfigure}%
        \vspace{-10pt}
        \caption{\small Throughput scaling when training GoogLeNet, VGG19 and VGG19-22K using Poseidon-parallelized Caffe with \emph{varying network bandwidth}. Single-node Caffe is set as baseline (speedup = 1). }
\label{fig:bandwidth}
\vspace{-18pt}
\end{figure*}

\begin{figure}[t]
\centering
        \begin{subfigure}[b]{0.235\textwidth}
                \includegraphics[width=\linewidth]{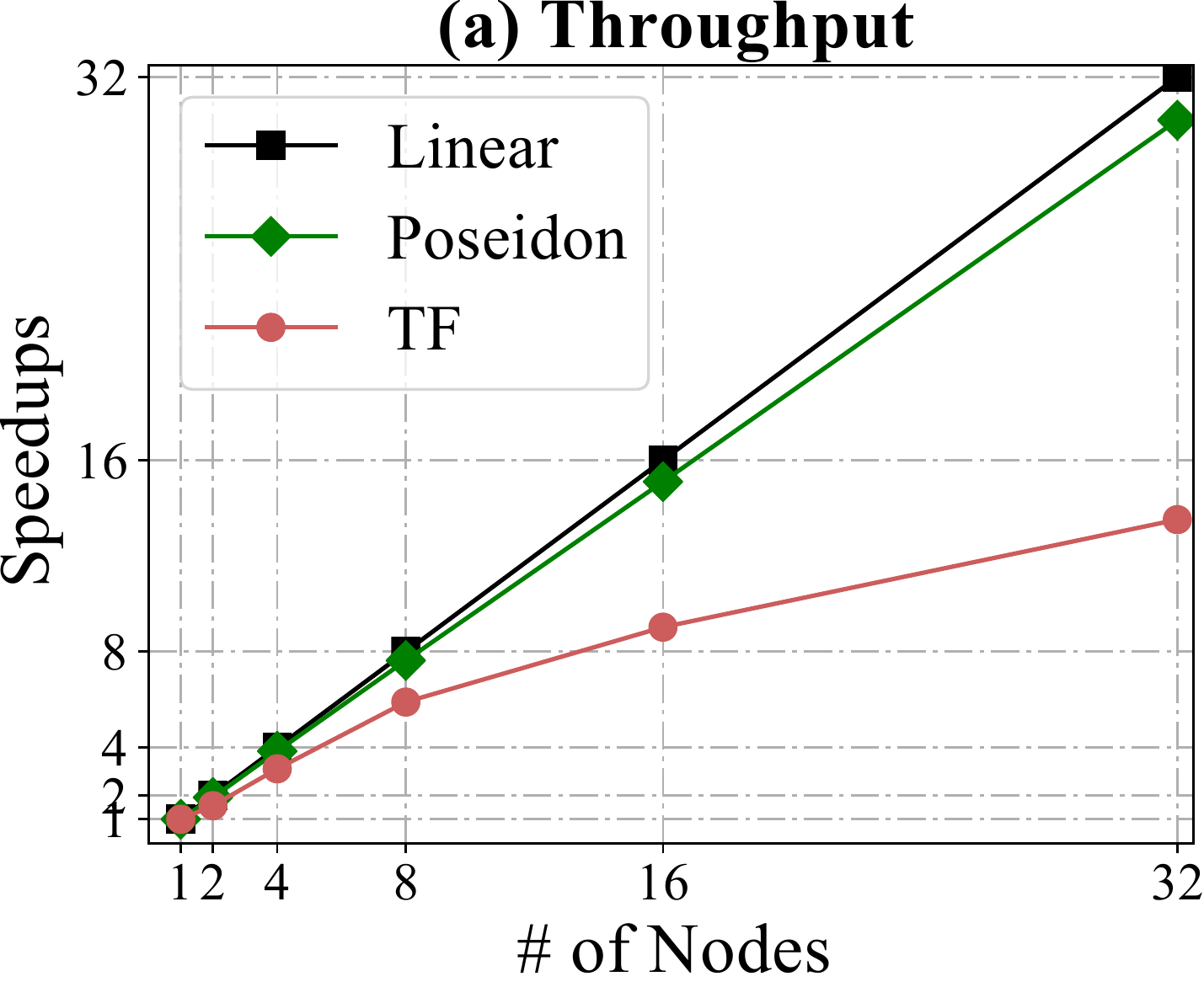}
        \end{subfigure}%
        \begin{subfigure}[b]{0.25\textwidth}
                \includegraphics[width=\linewidth]{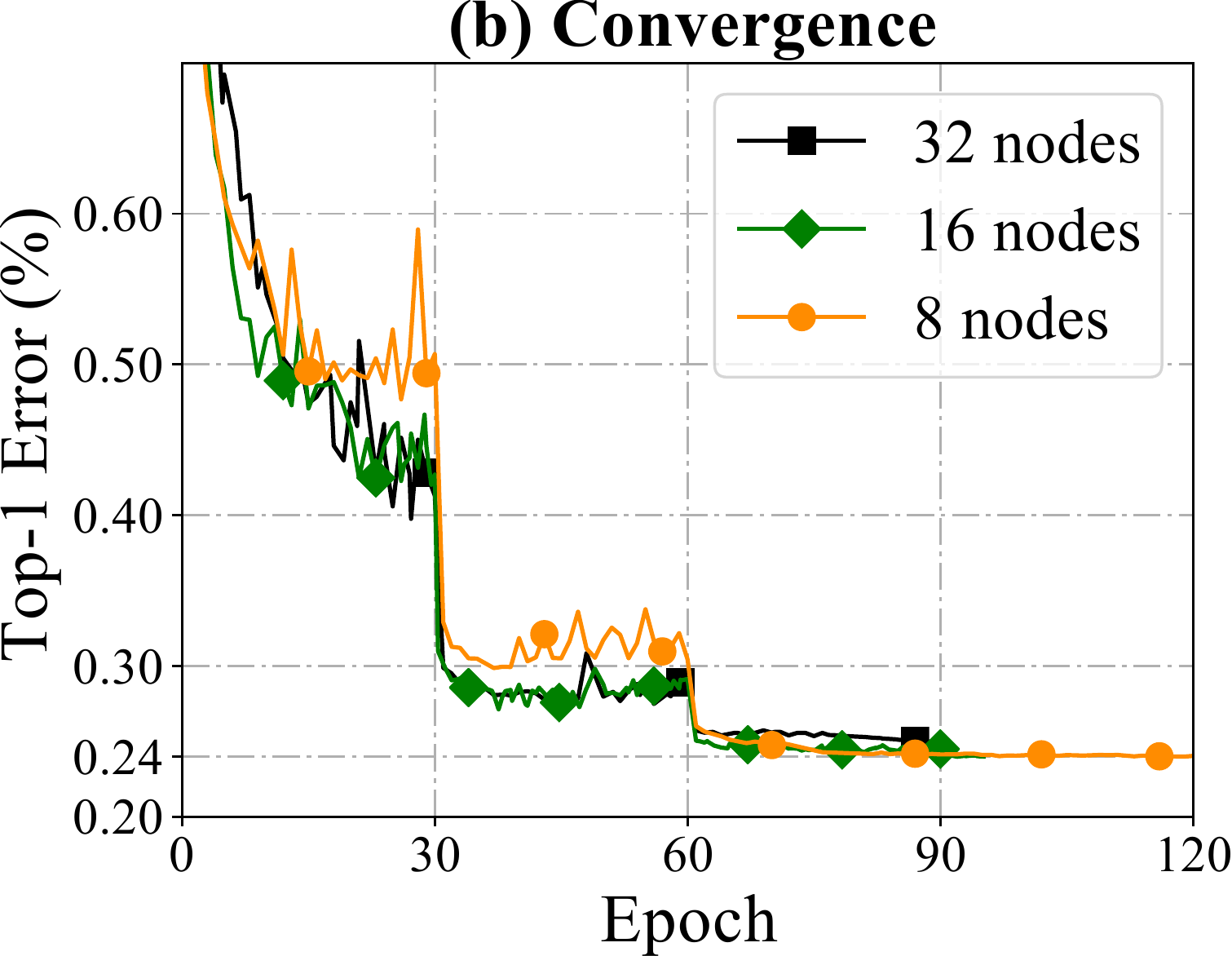}
        \end{subfigure}%
        \vspace{-8pt}
        \caption{\small (a) Speedup vs. number of nodes and (b) Top-1 test error vs. epochs for training ResNet-152 using Poseidon-TensorFlow and the original TensorFlow. }
\label{fig:resnet}
\end{figure}

\noindent \textbf{Statistical Performance.}
For completeness, we report in Figure~\ref{fig:resnet} the statistical performance for training ResNet-152 using Poseidon. Poseidon achieves near-linear speedups on both system throughput and statistical convergence: Poseidon delivers 31x speedup in terms of throughput, and reaches 0.24 reported error with less than 90 epochs with both 16 and 32 nodes -- thus linear scales in terms of time to accuracy, compared to 8 nodes with batchsize = $32 \times 8$, which is a standard setting as in~\cite{he2015deep}, echoing recent results that synchronous training on distributed GPUs yields better performance than asynchronous training in terms of time to quality for most NNs~\cite{cui2016geeps,chen2016revisiting}. For other NNs in Table.~\ref{tb:model_stats}, Poseidon delivers the same quality of accuracies as reported in their papers~\cite{Krizhevsky:2012:NIPS,szegedy2015rethinking,Szegedy:2014:going,Simonyan:2015:ICLR} on up to 32 GPUs.

\begin{figure}[bt]
\centering
\vspace{-12pt}
\includegraphics[width=0.85\columnwidth]{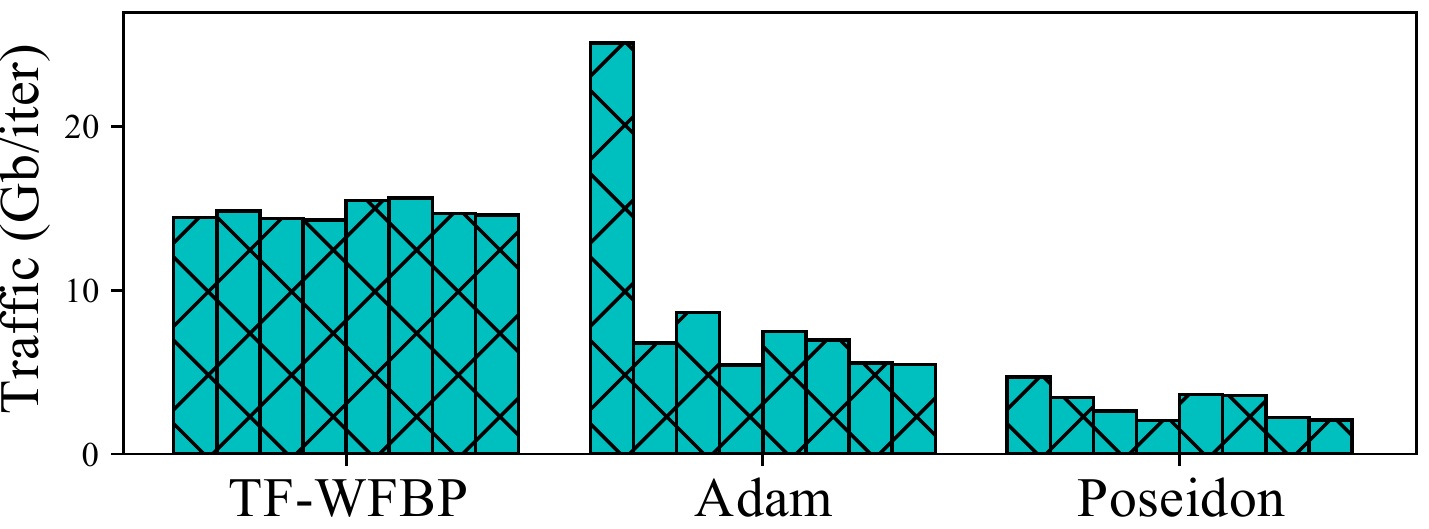}
\vspace{-10pt}
\caption{\small Averaged communication load when training VGG19 using \emph{TF-WFBP}, \emph{Adam} and \emph{Poseidon} with TensorFlow engine. Each bar represents the network traffic on a node.}
\label{fig:comm_load_tf}
\vspace{-15pt}
\end{figure}

\vspace{-12pt}
\subsection{Bandwidth Experiments}
\label{sec:evaluation:bandwidth}
\vspace{-5pt}
To further assess Poseidon's HybComm strategy, we simulate the environment where network bandwidth is limited. We use Linux traffic control tool \emph{tc} to lower the available bandwidth on each node, and compare the training throughput between with and without HybComm. We focus on Caffe engine in this section because it is lighter and less optimized than TensorFlow.

Figure~\ref{fig:bandwidth} plots the speedup on throughput vs. number of workers when training GoogLeNet, VGG19 and VGG19-22K with different maximum bandwidth. Clearly, limited bandwidth prevents a standard PS-based system from linearly scaling with number of nodes; for example, given 10GbE bandwidth (which is a commonly-deployed Ethernet configuration in most cloud computing platforms), training VGG19 using PS on 16 nodes can only be accelerated by 8x. This observation confirms our argument that limited bandwidth would result in communication bottleneck when training big models on distributed GPUs. Fortunately, Poseidon significantly alleviates this issue. Under limited bandwidth, it constantly improves the throughput by directly reducing the size of messages needed to be communicated, especially when the batch size is small; when training VGG19 and VGG19-22K, Poseidon achieves near-linear speedup on 16 machines using only 10GbE bandwidth, while an optimized PS would otherwise need 30GbE or even higher to achieve. Note that Poseidon will never underperform a traditional PS scheme because it will reduce to a parameter server whenever it results in less communication overheads; for instance, we observe that Poseidon reduces to PS when training GoogLeNet on 16 nodes, because GoogleNet only has one thin FC layer ($1000 \times 1024$) and is trained with a large batch size (128).

\vspace{-15pt}
\subsection{Comparisons to Other Methods}
\label{sec:evaluation:comparison_to_others}
\vspace{-5pt}
In this section, we compare Poseidon against other communication methods, including Adam~\cite{Chilimbi:2014:OSDI} and CNTK 1-bit quantization~\cite{yu2014introduction}, and show Poseidon's advantages.

\noindent \textbf{Adam.}
To save bandwidth, Adam~\cite{Chilimbi:2014:OSDI} synchronizes the parameters of a FC layer by first pushing SFs generated on all workers to a PS node, and then pulling back the full parameter matrices thereafter.
As direct comparisons to Adam~\cite{Chilimbi:2014:OSDI} are inaccessible, we implement its strategy in Poseidon, and compare it (denoted as \emph{Adam}) to \emph{TF-WFBP} and \emph{Poseidon} by monitoring the network traffic of each machine
when training VGG19 on 8 nodes using TensorFlow engine. As shown in Figure~\ref{fig:comm_load_tf}, the communication workload is highly imbalanced using Adam's strategy. Unlike a traditional PS (TF-WFBP) where the parameters are equally distributed over multiple shards, Adam cannot partition the parameters of FC layers because of their usage of SFs. Although the ``push'' operation uses SFs to reduce message size, the ``pull'' requires some server nodes to broadcast big matrices to each worker node, which creates bursty traffic that results in communication bottleneck on them.
By contrast, Poseidon either partitions parameters equally over multiple PS shards, or transmits SFs among peer workers, both are communication load-balanced that avoid bursty communication situations. Quantitatively, Adam delivers 5x speedup with 8 nodes when training VGG19.

\begin{figure}[tbh]
\centering
\vspace{-8pt}
        \begin{subfigure}[b]{0.24\textwidth}
                \includegraphics[width=\linewidth]{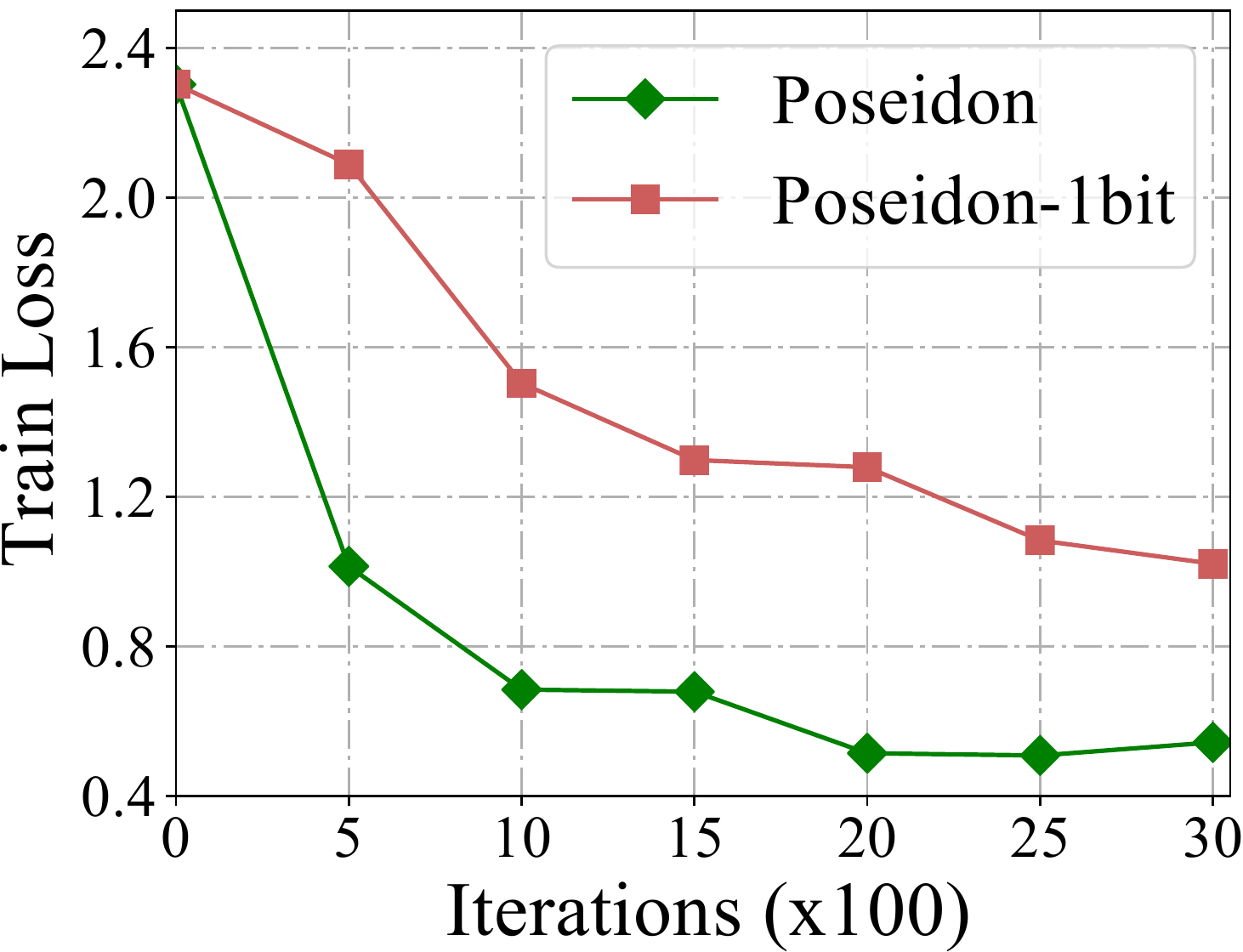}
        \end{subfigure}%
        \begin{subfigure}[b]{0.24\textwidth}
                \includegraphics[width=\linewidth]{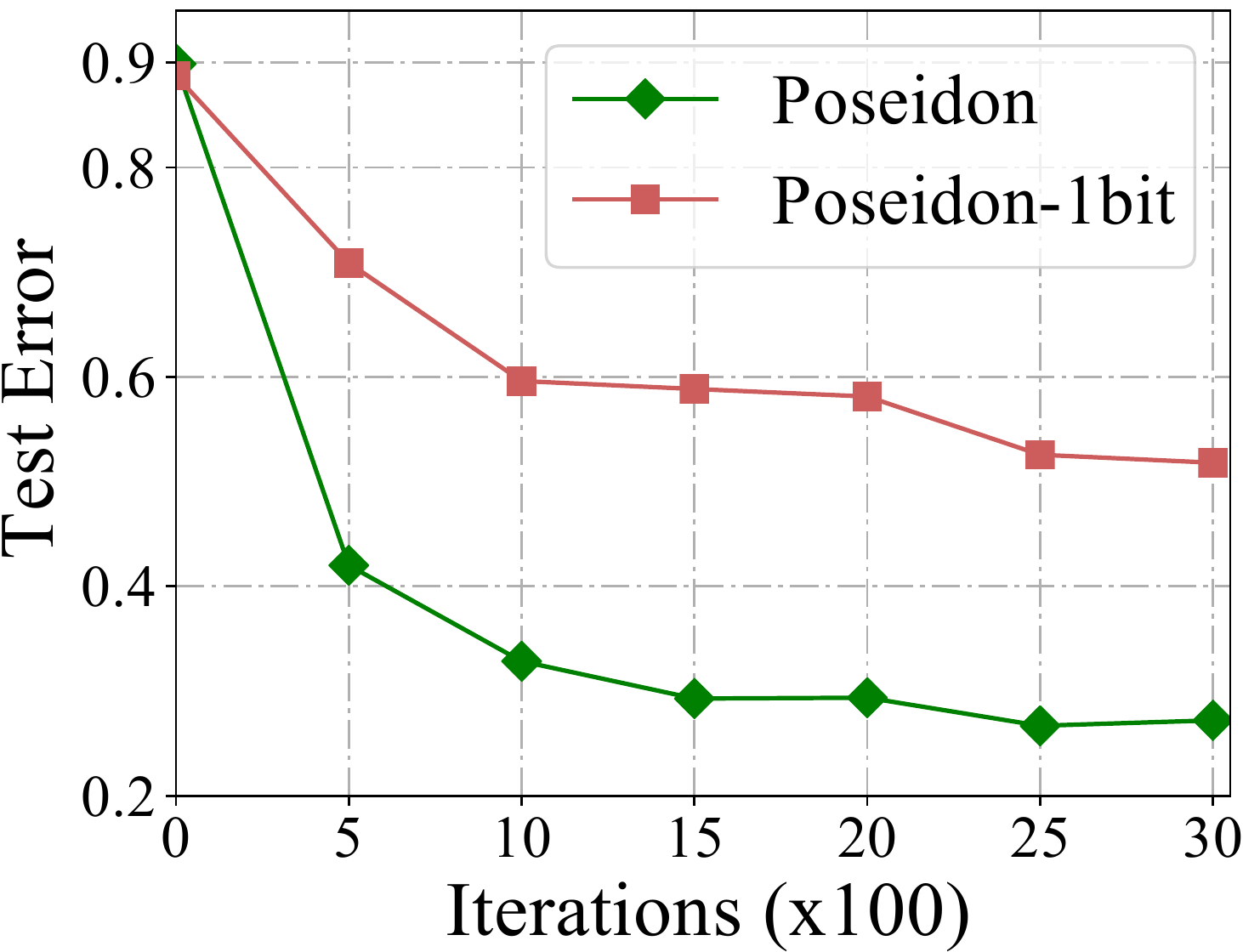}
        \end{subfigure}%
        \vspace{-8pt}
        \caption{\small Training loss and test error vs. iteration when training CIFAR-10 quick network using Poseidon and Poseidon-1bit on 4GPUs with Caffe engine. }
\label{fig:1bit}
\vspace{-20pt}
\end{figure}

\noindent \textbf{CNTK.}
We compare Poseidon to the 1-bit quantization technique proposed in CNTK~\cite{yu2014introduction}. We create a baseline \emph{Poseidon-1bit} which uses the 1-bit strategy to quantize the gradients in FC layers, and add the residual to updates of the next iteration. We then train the CIFAR-10 quick network, and plot the training loss and test error vs. iterations for two systems (both have linear scaling on throughput). As in Figure~\ref{fig:1bit}, 1-bit quantization yields worse convergence in terms of accuracy -- on 4 GPUs, it achieves 0.5 error after 3K iterations, while Poseidon quickly converges to 0.3 error at iteration 1000. We conjecture this is caused by the quantization residual, which is equivalent to delayed updates that may hurt the convergence performance when training NNs on images, confirmed by~\cite{cui2016geeps}. We also directly train VGG19 using CNTK-1bit system, and report 5.8x, 11x, 20x speedups on 8, 16 and 32 nodes, respectively, thus less scale-ups than Poseidon, and also compromised statistical performance due to approximated updates. 

\vspace{-5pt}
\section{Related Work}
\label{sec:relatedwork}
\vspace{-8pt}
\noindent \textbf{PS-based Distributed DL Systems.}
Based on the parameter server~\cite{Wei:2015:SoCC,li2014scaling} architecture, a number of CPU-based distributed DL systems have been developed, such as \cite{Zou:2014:VLDB,Wang:2015:MM,Dean:2012:NIPS,Le:2012:ICML} and Adam~\cite{Chilimbi:2014:OSDI}. 
They are purely PS-based systems on CPU-only clusters, whereas we address the more challenging case of GPU clusters.

Scaling up DL on distributed GPUs is an active field of research. Coates \etal\cite{Coates:2013:ICML} build a GPU-based multi-machine system for DL using model parallelism rather than data parallelism, and their implementation is rather specialized for a fixed model structure while demanding specialized hardware, such as InfiBand networking. TensorFlow~\cite{abadi2016tensorflow} is Google's distributed ML platform that uses a dataflow graph to represent DL models, and synchronizes model parameters via PS. It therefore cannot dynamically adjust its communication method depending on the layer and cluster information as Poseidon does. 
MXNet~\cite{chen2015mxnet} is another DL system that uses PS for distributed execution, and supports TensorFlow-like graph representations for DL models. By auto-parallelizing independent subgraphs, both frameworks implicitly overlap the communication and computation. By contrast, Poseidon has a more explicit way to overlap them via its client library. Hence, Poseidon can be also used to parallelize non-graph-based frameworks. Moreover, both MXNet and TensorFlow do not address the bottleneck caused by limited network bandwidth, which undermines their scalability when training large models with dense layers (e.g., big softmax).
Besides, Cui \etal propose GeePS~\cite{cui2016geeps} that manages the limited GPU memory and report speedups on distributed GPUs. While, GeePS does not address the issue of limited network bandwidth. Therefore, Poseidon's technique could be combined with them to enable better training speedups.
Also of note are several efforts to port Caffe onto other distributed platforms, such as SparkNet~\cite{moritz2015sparknet}, YahooCaffe~\cite{YahooCaffe} and FireCaffe~\cite{iandola2016firecaffe}, the former reports a 4-5 times speedup with 10 machines (and hence less scalability than our results herein).

\noindent \textbf{Other distributed ML systems.}
CNTK~\cite{yu2014introduction} is a DL framework that supports distributed executions and addresses the problem of communication bottleneck via the 1-bit quantization technique.
CNTK demonstrates little negative impact on convergence in speech domains~\cite{seide2014parallelizability,seide20141}. However, in some other domains (Section \ref{sec:evaluation:comparison_to_others}), the performance is usually compromised by noisy gradients~\cite{abadi2016tensorflow,cui2016geeps}. By contrast, Poseidon's HybComm reduces the communication while always guaranteeing synchronous training. 
There are also growing interest in parallelizing ML applications using peer-to-peer communication, such as MALT~\cite{li2015malt}, SFB~\cite{Xie:2015:arXiv} and Ako~\cite{watcharapichat2016ako}.
Poseidon draws inspiration from these works
but goes one step further as it is an adaptive best-of-both-worlds protocol, which will select client-server communication whenever it would result in fewer overheads. 


\vspace{-15pt}
\section{Conclusion}
\label{sec:conclusion}
\vspace{-8pt}
We present Poseidon, a scalable and efficient communication architecture for large-scale DL on distributed GPUs. 
Poseidon's design is orthogonal to TensorFlow, Caffe or other DL frameworks -- the techniques present in Poseidon could be used to produce a better distributed version of them.
We empirically show that Poseidon constantly delivers linear speedups using up to 32 nodes and limited bandwidth on a variety of neural network, datasets and computation engines, and compares favorably to Adam and Microsoft CNTK. 

\vspace{-15pt}
\section*{Acknowledgments}
\vspace{-10pt}
We thank our shepherd Yu Hua and ATC reviewers for their helpful feedback. We thank the CMU Parallel Data Laboratory for their machine resources and Henggang Cui for insightful discussion. This research is supported by NSF Big Data IIS1447676 and NSF XPS Parallel CCF1629559.

{\bibliographystyle{acm}
\bibliography{poseidon}}


\end{document}